\documentclass[10pt,twocolumn,letterpaper]{article}

\usepackage{iccv}
\usepackage{times}
\usepackage{epsfig}
\usepackage{graphicx}
\usepackage{amsmath}
\usepackage{amssymb}

\usepackage{nicefrac}       % compact symbols for 1/2, etc.
\usepackage{microtype}      % microtypography

\usepackage{multirow}
\usepackage{verbatim}
\usepackage{color}

\usepackage{multicol}

\usepackage{blindtext}
\usepackage{float}
\usepackage{enumitem}
\usepackage{booktabs}
\usepackage{tabulary,multirow,overpic,xcolor}

\usepackage{algorithm}
\usepackage{listings}

\usepackage[caption=false]{subfig}
\usepackage{pifont}% http://ctan.org/pkg/pifont
\usepackage{epstopdf}
\usepackage{algorithm}
\usepackage{algorithmicx}
\usepackage{algpseudocode}
\usepackage{enumerate}

\usepackage{bm}
\usepackage{t1enc}
\usepackage{colortbl}
\usepackage[nosort]{cite}

\usepackage[british,UKenglish,USenglish,english,american]{babel}

\newcommand{\figref}[1]{Fig.~\ref{#1}}
\newcommand{\tblref}[1]{Table~\ref{#1}}
\newcommand{\sref}[1]{Sec.~\ref{#1}}

\usepackage{soul}
\definecolor{carmine}{rgb}{0.59, 0.0, 0.09}

\newcommand{\app}{\raise.17ex\hbox{$\scriptstyle\sim$}}

\def\x{$\times$}

\newcommand{\fullnameCC}{Multiscale Transformer\xspace}
\newcommand{\attnnameCC}{Multi Head Pooling Attention\xspace}
\newcommand{\attnname}{multi head pooling attention\xspace}
\newcommand{\attnabbvspace}{MHPA\xspace}
\newcommand{\singleattnname}{pooling attention\xspace}
\newcommand{\attnabbv}{MHPA\xspace}

\newcommand{\pacc}[1]{{\bf \fontsize{7.5}{42}\selectfont \color{citecolor!80}~(#1)}}
\newcommand{\macc}[1]{{\bf \fontsize{7.5}{42}\selectfont \color{lightred!180}~(#1)}}

\newcolumntype{x}[1]{>{\centering\arraybackslash}p{#1pt}}

\newlength\savewidth\newcommand\shline{\noalign{\global\savewidth\arrayrulewidth
		\global\arrayrulewidth 1pt}\hline\noalign{\global\arrayrulewidth\savewidth}}
\newcommand{\tablestyle}[2]{\setlength{\tabcolsep}{#1}\renewcommand{\arraystretch}{#2}\centering\footnotesize}
\makeatletter\renewcommand\paragraph{\@startsection{paragraph}{4}{\z@}
	{.5em \@plus1ex \@minus.2ex}{-.5em}{\normalfont\normalsize\bfseries}}\makeatother

\definecolor{citecolor}{RGB}{34,139,34}
\definecolor{citecolor2}{HTML}{0071bc}
\definecolor{lightred}{RGB}{241,140,142}
\usepackage[pagebackref=true,breaklinks=true,letterpaper=true,colorlinks,
citecolor=citecolor2,bookmarks=false]{hyperref}

 \iccvfinalcopy % *** Uncomment this line for the final submission

\def\iccvPaperID{***} % *** Enter the ICCV Paper ID here

\begin{document}
	
	%%%%%%%%% TITLE
	\title{ 
	\vspace{-.5em} Multiscale Vision Transformers  \\ \vspace{-.7em}
	}
\author{
	Haoqi Fan\textsuperscript{ *, 1} \qquad
	Bo Xiong\textsuperscript{ *, 1} \qquad
	Karttikeya Mangalam\textsuperscript{ *, 1, 2}  \vspace{.1em} \\
	Yanghao Li\textsuperscript{ *, 1}   \qquad
	Zhicheng Yan\textsuperscript{ 1} \qquad \quad
	Jitendra Malik\textsuperscript{ 1, 2} \qquad \quad
	Christoph Feichtenhofer\textsuperscript{ *, 1}  \vspace{.8em}\\
	\textsuperscript{1}Facebook AI Research \qquad \qquad \textsuperscript{2}UC Berkeley 
	 \vspace{-.8em}
}
\maketitle
\ificcvfinal
    \renewcommand*{\thefootnote}{\fnsymbol{footnote}}
    \setcounter{footnote}{1}
    \footnotetext{Equal technical contribution.}
    \renewcommand*{\thefootnote}{\arabic{footnote}}
    \setcounter{footnote}{0}
\fi

	
\definecolor{fastcolor}{RGB}{100,178,100}
\definecolor{slowcolor}{RGB}{120,120,243}
\definecolor{expandcolor}{RGB}{244,157,78}
\definecolor{clipscolor}{RGB}{110,216,230}
\newcommand{\fastcolor}[1]{\textcolor{fastcolor}{#1}}
\newcommand{\fastcolorC}[1]{\textcolor{orange}{#1}}
\newcommand{\slowcolor}[1]{\textcolor{slowcolor}{#1}}
\newcommand{\expandcolor}[1]{\textcolor{expandcolor}{#1}}
\newcommand{\clipscolor}[1]{\textcolor{clipscolor}{#1}}

\newcommand{\slow}{\slowcolor{Slow }}
\newcommand{\fast}{\fastcolor{Fast }}

\definecolor{predictioncolor}{RGB}{0,255,0}
\definecolor{labelcolor}{RGB}{255,0,0}
\newcommand{\predictioncolor}[1]{\textcolor{predictioncolor}{#1}}
\newcommand{\labelcolor}[1]{\textcolor{labelcolor}{#1}}

\newcommand{\pred}{\predictioncolor{\textbf{Predictions}: }}
\newcommand{\gt}{\labelcolor{\textbf{Labels}: }}

\definecolor{demphcolor}{RGB}{144,144,144}
\newcommand{\demph}[1]{\textcolor{demphcolor}{#1}}

\definecolor{xycolor}{RGB}{60, 120, 216}
\definecolor{xycolor}{HTML}{0071bc}
\newcommand{\xycolor}[1]{\textcolor{xycolor}{#1}}
\definecolor{wcolor}{RGB}{103, 78, 167}
\newcommand{\wcolor}[1]{\textcolor{wcolor}{#1}}
\definecolor{dcolor}{RGB}{166, 77,21}
\newcommand{\dcolor}[1]{\textcolor{dcolor}{#1}}
\definecolor{gcolor}{RGB}{204, 102, 153}
\newcommand{\gcolor}[1]{\textcolor{gcolor}{#1}}
\definecolor{tcolor}{RGB}{80, 200, 180}
\newcommand{\tcolor}[1]{\textcolor{citecolor}{#1}}
\definecolor{eicolor}{RGB}{153, 51, 102}
\newcommand{\eicolor}[1]{\textcolor{eicolor}{#1}}

\def\gab{\textcolor{eicolor}{ $\bm{\gamma_{b}}$}}
\def\gaw{\textcolor{wcolor}{ $\bm{\gamma_w$}}}
\def\gag{\textcolor{gcolor}{ $\bm{\gamma_{g}$}}}
\def\gax{\textcolor{xycolor}{$\bm{\gamma_x$}}}
\def\gay{\textcolor{xycolor}{$\bm{\gamma_y$}}}
\def\gaxy{\textcolor{xycolor}{$\bm{\gamma_{s}$}}}
\def\gat{\textcolor{tcolor}{$\bm{\gamma_t$}}}
\def\gatau{\textcolor{fastcolor}{$\bm{\gamma_\tau$}}}
\def\gabeta{\textcolor{orange}{$\bm{\gamma_\beta$}}}
\def\gaalpha{\textcolor{fastcolor}{$\bm{\gamma_\alpha$}}}
\def\gad{\textcolor{dcolor}{$\bm{\gamma_d$}}}

%##################################################################################################
\vspace*{-1.0em}
\begin{abstract}
\vspace*{-0.8em}
We present Multiscale Vision Transformers (MViT) for video and image recognition, by connecting the seminal idea of multiscale feature hierarchies with transformer models. Multiscale Transformers have several channel-resolution scale stages. Starting from the input resolution and a small channel dimension, the stages hierarchically expand the channel capacity while reducing the spatial resolution. This creates a multiscale pyramid of features with  early layers operating at high spatial resolution to model simple low-level visual information, and deeper layers at spatially coarse, but complex, high-dimensional features. We evaluate this fundamental architectural prior for modeling the dense nature of visual signals for a variety of video recognition tasks where it outperforms concurrent vision transformers that rely on large scale external pre-training and are 5-10\x~more costly in computation and parameters. We further remove the temporal dimension and apply our model for image classification where it  outperforms prior work on vision transformers. Code is available at: \url{https://github.com/facebookresearch/SlowFast}. 
\end{abstract}

\vspace{-5pt}	
\section{Introduction}
\vspace{-5pt}	

\label{sec:introduction}
We begin with the intellectual history of neural network models for computer vision. Based on their studies of cat and monkey visual cortex, Hubel and Wiesel \cite{hubel1960receptive} developed a \textit{hierarchical} model of the visual pathway with neurons in lower areas such as V1 responding to features such as oriented edges and bars, and in higher areas to more specific stimuli. Fukushima proposed the Neocognitron \cite{fukushima1982neocognitron}, a neural network architecture for pattern recognition explicitly motivated by Hubel and Wiesel’s hierarchy. His model had alternating layers of simple cells and complex cells, thus incorporating downsampling, and shift invariance, thus incorporating convolutional structure. LeCun \etal \cite{lecun1989backpropagation} took the additional step of using backpropagation to train the weights of this network. But already the main aspects of hierarchy of visual processing had been established: (\textit{i}) Reduction in spatial resolution as one goes up the processing hierarchy and (\textit{ii}) Increase in the number of different ``channels'', with each channel corresponding to ever more specialized features.

In a parallel development, the computer vision community developed \textit{multiscale} processing, sometimes called “pyramid” strategies, with Rosenfeld and Thurston \cite{rosenfeld1971edge}, Burt and Adelson \cite{burt1987laplacian}, Koenderink \cite{koenderink1984structure}, among the key papers. There were two motivations (\textit{i}) To decrease the computing requirements by working at lower resolutions and (\textit{ii})~A better sense of “context” at the lower resolutions, which could then guide the processing at higher resolutions (this is a precursor to the benefit of “depth” in today’s neural networks.)

The Transformer~\cite{vaswani2017attention} architecture 
allows learning arbitrary functions defined over sets and has been scalably successful in sequence tasks such as language comprehension \cite{edunov2018understanding} and machine translation \cite{brown2020language}. Fundamentally, a transformer uses blocks with two basic operations. First, is an attention operation \cite{bahdanau2014neural} for modeling inter-element relations. Second, is a multi-layer perceptron (MLP), which models relations within an element. Intertwining these operations with normalization \cite{ba2016layer} and residual connections \cite{He2015} 
allows transformers to generalize to a wide variety of tasks.

\begin{figure}
    \centering
    \includegraphics[width = 0.99\columnwidth]{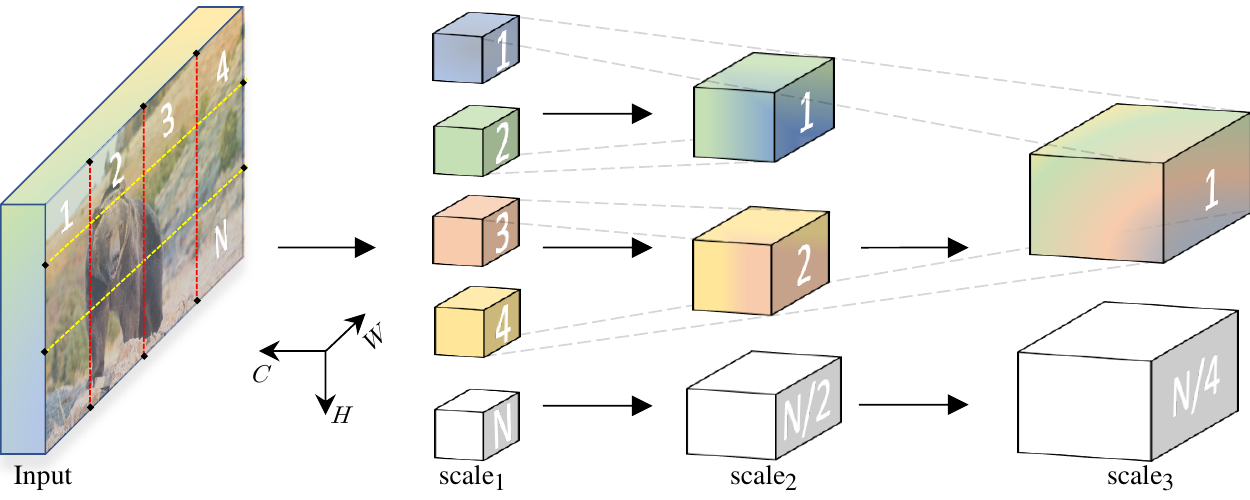}
        \vspace{-1pt}
    \caption{\textbf{Multiscale Vision Transformers} learn a hierarchy from \textit{dense} (in space) and \textit{simple} (in channels)  to \textit{coarse} and \textit{complex} features. 
    Several resolution-channel \textit{scale} stages progressively \textit{increase} the channel capacity of the intermediate latent sequence while \textit{reducing} its length and thereby spatial resolution.}
    \label{fig:teaser}
    \vspace{-10pt}
\end{figure}

Recently, transformers have been applied to key computer vision tasks such as image classification. In the spirit of architectural universalism, vision transformers \cite{dosovitskiy2020image,touvron2020training} approach performance of convolutional models across a variety of data and compute regimes. By only having a first layer that `patchifies' the input in spirit of a 2D convolution, followed by a stack of transformer blocks, the vision transformer aims to showcase the power of the transformer architecture using little inductive bias.

In this paper, our intention is to connect the seminal idea of \textit{multiscale feature hierarchies} with the transformer model. We posit that the fundamental vision principle of resolution and channel scaling,  
can be beneficial for transformer models across a variety of visual recognition tasks. 

We present Multiscale Vision Transformers (MViT), a transformer architecture for modeling visual data such as images and videos. Consider an input image as shown in~\figref{fig:teaser}. Unlike conventional transformers, which maintain a constant channel capacity and resolution throughout the network, Multiscale Transformers have several channel-resolution `scale' stages. Starting from the image resolution and a small channel dimension, the stages \textit{ hierarchically expand} the \textit{channel} capacity while \textit{reducing} the \textit{spatial} resolution. This creates a multiscale pyramid of feature activations inside the transformer network, effectively connecting the principles of transformers with multi scale feature hierarchies. 

Our conceptual idea provides an effective design advantage for vision transformer models. The early layers of our architecture can operate at high spatial resolution to model \textit{simple} low-level visual information, due to the lightweight channel capacity. In turn, the deeper layers can effectively focus on spatially coarse but \textit{complex} high-level features to model visual semantics. The fundamental advantage of our multiscale transformer arises from the extremely dense nature of  visual signals, a phenomenon that is even more pronounced for space-time visual signals captured in \textit{video}. 

A noteworthy benefit of our design is the presence of strong implicit temporal bias in video multiscale models. We show that vision transformer models~\cite{dosovitskiy2020image} trained on natural video suffer no performance decay when tested on videos with \textit{shuffled} frames. This indicates that these models are not effectively using the temporal information and instead rely heavily on appearance. In contrast, when testing our MViT models on shuffled frames,  we observe significant accuracy decay, indicating strong use of temporal information. 

Our focus in this paper is video recognition, and we design and evaluate MViT for video tasks (Kinetics~\cite{Kay2017,Carreira2018}, Charades \cite{Sigurdsson2016}, SSv2~\cite{ssv2} and AVA \cite{Gu2018}). MViT provides a significant performance gain over concurrent video transformers~\cite{neimark2021video,bertasius2021space,arnab2021vivit}, \textit{without} any external pre-training data. 

\begin{figure}[t]
	\centering
			\vspace{-0.8em}
	\includegraphics[width=1.0\linewidth]{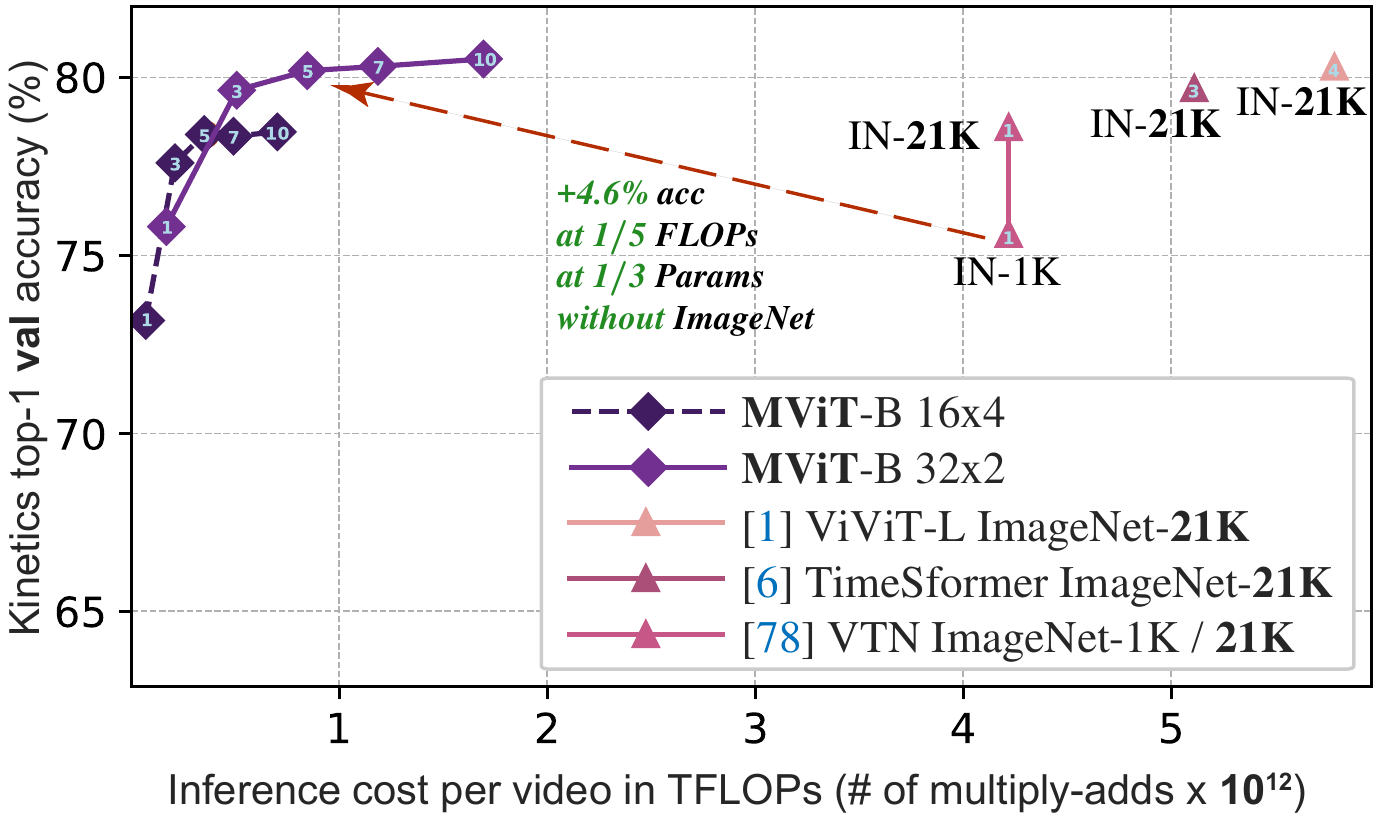}
	\caption{\textbf{Accuracy/complexity trade-off} on Kinetics-400 for varying  \# of inference clips per video shown in MViT curves. 
		Concurrent vision-transformer based methods~\cite{neimark2021video,bertasius2021space,arnab2021vivit} require over 5\x~more computation \textit{and  large-scale external pre-training} on ImageNet-\textbf{21K} (IN-\textbf{21K}), to achieve equivalent \textbf{MViT} accuracy. 
	}
	\label{fig:inference_cost}
		\vspace{-0.8em}
\end{figure}

 In \figref{fig:10clipTestAll} we show the computation/accuracy trade-off for video-level inference, when varying the number of temporal clips used in MViT. The vertical axis shows  accuracy on Kinetics-400  and the horizontal axis the overall inference cost in FLOPs for different  models, \textbf{MViT} and concurrent ViT~\cite{dosovitskiy2020image} video variants: VTN \cite{neimark2021video},  TimeSformer \cite{bertasius2021space}, ViViT \cite{arnab2021vivit}. To achieve similar accuracy level as MViT, these models require significant more computation and parameters (\eg ViViT-L~\cite{arnab2021vivit} has 6.8\x~higher  FLOPs and 8.5\x~more parameters at equal accuracy, more analysis in \S\ref{sec:ablations_kinetics}) and need large-scale external pre-training on ImageNet-\textbf{21K} (which contains around 60\x~more labels than Kinetics-400). 

We further apply our architecture to an image classification task on \mbox{ImageNet}~\cite{deng2009imagenet}, by \textit{simply removing the temporal dimension} of the video model found with ablation experiments on Kinetics, and show significant gains over single-scale vision transformers for image recognition.

\section{Related Work}

\noindent\textbf{Convolutional networks (ConvNets)}. Incorporating downsampling, shift invariance, and shared weights, ConvNets are de-facto standard backbones for computer vision  tasks for image~\cite{lecun1989backpropagation,Krizhevsky2012,Simonyan2015,Szegedy2015,He2016a, chen2018big, chen2019drop, gao2019res2net, tan2019efficientnet, ilija_2020, resnest}  and video~\cite{Simonyan2014,Feichtenhofer2016, Carreira2017, Qiu2017, Li2018, Xie2018,Tran2019, Feichtenhofer2019, Wu2019, girdhar2019video, feichtenhofer2020x3d, Zhou2017, jiang2019stm}. 

\noindent\textbf{Self-attention in ConvNets.}
Self-attention mechanisms has been used for
image understanding \cite{ramachandran2019stand, zhao2020exploring, hu2019local}, unsupervised object recognition \cite{locatello2020object} as well as vision and language~\cite{lu2019vilbert, li2019visualbert}. 
Hybrids of self-attention operations and convolutional networks have also  been
applied to 
image understanding \cite{hu2018relation} and video recognition \cite{Wang2018}.

\noindent\textbf{Vision Transformers.} Much of current enthusiasm in application of Transformers \cite{vaswani2017attention} to vision tasks commences with the Vision Transformer (ViT) \cite{dosovitskiy2020image} and Detection Transformer  \cite{carion2020end}. We build directly upon \cite{dosovitskiy2020image} with a staged model allowing channel expansion and resolution downsampling. DeiT \cite{touvron2020training} proposes a data efficient approach to training ViT. Our training recipe builds on, and we compare our image classification models to, DeiT under identical settings.

An emerging thread of work aims at applying transformers to vision tasks such as object detection \cite{beal2020toward}, semantic segmentation \cite{zhao2020point,wang2020max}, 3D reconstruction \cite{lin2020end}, pose estimation \cite{yang2020transpose}, generative modeling \cite{chen2020generative}, image retrieval \cite{el2021training}, medical image segmentation \cite{chen2021transunet, valanarasu2021medical, yun2021spectr}, point clouds \cite{guo2020pct}, video instance segmentation \cite{wang2020end}, object re-identification \cite{he2021transreid}, video retrieval \cite{gabeur2020multi}, video dialogue \cite{le2019multimodal}, video object detection \cite{yuan2020temporal} and multi-modal tasks \cite{liu2021cptr, desai2020virtex, radford2021learning, hu2021transformer, yu2019multimodal}.
A separate line of works attempts at modeling visual data with learnt discretized token sequences \cite{wu2020visual, ramesh2021zero, chen2020generative,yuan2021tokens,chu2021we}. 

\noindent\textbf{Efficient Transformers.} Recent works \cite{wang2020linformer, kitaev2020reformer, choromanski2020rethinking, tay2020sparse, dai2020funnel, child2019generating, li2020linear} reduce the quadratic attention complexity to make transformers more efficient for natural language processing applications, which is complementary to our approach.

Three concurrent works propose a ViT-based architecture for video \cite{neimark2021video, bertasius2021space,arnab2021vivit}. However, these methods rely on pre-training on vast amount of external data such as ImageNet-21K \cite{deng2009imagenet}, and thus use the vanilla ViT~\cite{dosovitskiy2020image} with minimal adaptations. In contrast, our MViT introduces multiscale feature hierarchies for transformers, allowing effective modeling of dense visual input without large-scale external data. 

\label{sec:related_work}

%##################################################################################################
\newcommand{\blocks}[3]{\multirow{3}{*}{\(\left[\begin{array}{c}\text{1$\times$1$^\text{2}$, #2}\\[-.1em] \text{1$\times$3$^\text{2}$, #2}\\[-.1em] \text{1$\times$1$^\text{2}$, #1}\end{array}\right]\)$\times$#3}
}
\newcommand{\blocket}[4]{\multirow{3}{*}{\(\left[\begin{array}{c}\text{1$\times$1$^\text{2}$, #1}\\[-.1em] \text{$3$$\times$3$^\text{2}$, #2}\\[-.1em] \text{1$\times$1$^\text{2}$, #3}\end{array}\right]\)$\times$#4}
}
\newcommand{\blockatt}[3]{\multirow{2}{*}{\(\left[\begin{array}{c}\text{MHA(\wcolor{#1})}\\[-.1em] \text{MLP(\wcolor{#2})}\end{array}\right]\)$\times$#3}
}
\newcommand{\blockatta}[3]{\multirow{2}{*}{\(\left[\begin{array}{c}\text{\eicolor{MHPA}(\wcolor{#1})}\\[-.1em] \text{MLP(\wcolor{#2})}\end{array}\right]\)$\times$#3}
}

\newcommand{\blockt}[3]{\multirow{3}{*}{\(\left[\begin{array}{c}\text{\underline{3$\times$1$^\text{2}$}, #2}\\[-.1em] \text{1$\times$3$^\text{2}$, #2}\\[-.1em] \text{1$\times$1$^\text{2}$, #1}\end{array}\right]\)$\times$#3}
}
\newcommand{\outsizes}[7]{\multirow{#7}{*}{\(\begin{array}{c} \text{\emph{Slow}}: \text{#1$\times$#2$^\text{2}$}\\[-.1em] \text{\emph{Fast}}: \text{#4$\times$#5$^\text{2}$}\end{array}\)}
}

\newcommand{\outsizesRaw}[4]{\multirow{#4}{*}{\(\begin{array}{c}  \text{#1$\times$#2\x #3}\\[-.1em]  \end{array}\)}}
\newcommand{\outsizesRawD}[5]{\multirow{#5}{*}{\(\begin{array}{c}  \text{#1\x#2\x#3\x#4}\\[-.1em]  \end{array}\)}}

\newcommand{\outsizesGamma}[4]{\multirow{#4}{*}{\(\begin{array}{c}  \text{#1\gat\x (#2\gaxy)$^2$}\\[-.1em]  \end{array}\)}}

\newcommand{\outsizesSF}[5]{\multirow{#5}{*}{\(\begin{array}{cc} \text{\emph{Slow}}:& \text{#1$\times$ #2$^\text{2}$}\\[-.1em] \text{\emph{Fast}}:& \text{#3$\times$#4$^\text{2}$}\end{array}\)}}

\section{Multiscale Vision Transformer (MViT)} \label{sec:method} 
Our generic Multiscale Transformer architecture builds on the core concept of \textit{stages}. Each stage consists of multiple transformer blocks with specific space-time resolution and channel dimension. The main idea of Multiscale Transformers is to progressively \textit{expand} the channel capacity, while \textit{pooling} the resolution from input to output of the network. 

\subsection{\attnnameCC}
\label{sec:pooling_attn}

\begin{figure}[t!]
	\centering
	\vspace{-5pt}
	\includegraphics[width = 1.0\columnwidth]{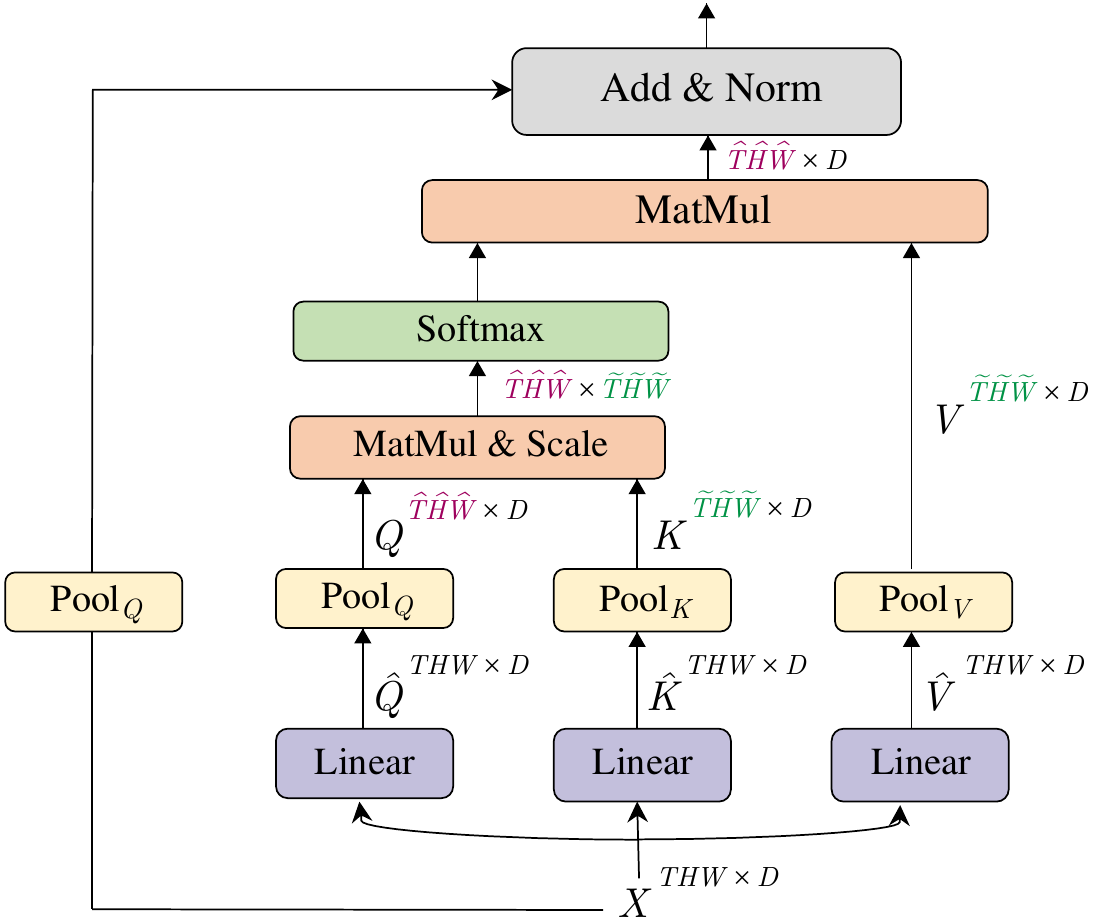}
	\caption{\textbf{Pooling Attention} is a flexible attention mechanism that (i) 
	allows obtaining the reduced space-time resolution ($\hat{T}\hat{H}\hat{W}$) of the input ($THW$)  by pooling the query,  $Q = \mathcal{P}(\hat{Q};
	{\mathbf{\Theta}_Q})$, and/or (ii) computes attention on a reduced length ($\tilde{T}\tilde{H}\tilde{W}$) by pooling the key, $K = \mathcal{P}(\hat{K}; {\mathbf{\Theta}_K})$, and value, $V = \mathcal{P}(\hat{V}; {\mathbf{\Theta}_V})$, sequences. }
	\label{fig:pooling_attn}
\end{figure}

We first describe \attnnameCC (\attnabbv), a self attention operator that enables flexible resolution modeling in a transformer block allowing Multiscale Transformers to operate at progressively changing spatiotemporal resolution. In contrast to original Multi Head Attention (MHA) operators \cite{vaswani2017attention}, where the channel dimension and the spatio-temporal resolution remains fixed, \attnabbvspace \textit{pools} the sequence of latent tensors to reduce the sequence length (resolution) of the attended input.  \figref{fig:pooling_attn} shows the concept.

Concretely, consider a $D$ dimensional input tensor $X$ of sequence length $L$, $X \in \mathbb{R}^{L \times D}$. Following MHA \cite{dosovitskiy2020image}, \attnabbv projects the input $X$ into intermediate query tensor $\hat{Q} \in \mathbb{R}^{L \times D}$, key tensor $\hat{K} \in \mathbb{R}^{L \times D}$ and value tensor $\hat{V} \in \mathbb{R}^{L \times D}$  with linear operations
$$
\hat{Q} = XW_{Q} \quad \hat{K} = XW_{K} \quad \hat{V} = XW_{V}
$$/
with weights $W_{Q},W_{K},W_{V}$ of dimensions $D \times D$. The obtained intermediate tensors are then pooled in sequence length, with a pooling operator $\mathcal{P}$ as described below. 

\paragraph{Pooling Operator.}
\label{subsec:pooling_op}
Before attending the input, the intermediate tensors $\hat{Q}, \hat{K}, \hat{V}$ are pooled with the pooling operator $\mathcal{P}(\cdot; \mathbf{\Theta})$ which is the cornerstone of our \attnabbv and, by extension, of our \fullnameCC architecture.

 The operator $\mathcal{P}(\cdot ; \mathbf{\Theta})$ performs a pooling kernel computation on the input tensor along each of the dimensions. Unpacking $\mathbf{\Theta}$ as $\mathbf{\Theta} := (\mathbf{k}, \mathbf{s}, \mathbf{p})$, the operator employs a pooling kernel $\mathbf{k}$ of dimensions $k_T \times k_H \times k_W$, a stride $\mathbf{s}$ of corresponding dimensions $s_T \times s_H \times s_W$ and a padding $\mathbf{p}$ of corresponding dimensions $p_T \times p_H \times p_W$ to reduce an input tensor of dimensions $\mathbf{L} = T \times H \times W$ to $\mathbf{\Tilde{L}}$ given by, 
$$
\mathbf{\Tilde{L}} = \left\lfloor \frac{\mathbf{L} + 2\mathbf{p} - \mathbf{k}}{\mathbf{s}} \right\rfloor + 1
$$
with the equation applying coordinate-wise. The pooled tensor is flattened again yielding the output of $\mathcal{P}(Y; \mathbf{\Theta}) \in \mathbb{R}^{\Tilde{L} \times D}$ with reduced sequence length, $\Tilde{L} = \Tilde{T} \times \Tilde{H} \times \Tilde{W}$. 

By default we use \textit{overlapping} kernels $\mathbf{k}$ with \textit{shape-preserving} padding $\mathbf{p}$ in our pooling attention operators, so that $\Tilde{L}$ , the sequence length of the output tensor $\mathcal{P}(Y ; \mathbf{\Theta})$, experiences an overall reduction by a factor of $s_Ts_Hs_W$.

\paragraph{Pooling Attention.}
 The pooling operator $P(\cdot; \Theta)$ is applied to all the intermediate tensors $\hat{Q}$, $\hat{K}$ and $\hat{V}$ independently with chosen pooling kernels $\mathbf{k}$, stride $\mathbf{s}$ and padding $\mathbf{p}$. Denoting $\theta$ yielding the pre-attention vectors $Q = \mathcal{P}(\hat{Q}; {\mathbf{\Theta}_Q})$, $K = \mathcal{P}(\hat{K}; {\mathbf{\Theta}_K})$ and $V = \mathcal{P}(\hat{V}; {\mathbf{\Theta}_V})$ with reduced sequence lengths. Attention is now computed on these shortened vectors, with the operation,
$$
\operatorname{Attention}(Q, K, V) = \operatorname{Softmax}(QK^T/\sqrt{D})V.
$$

Naturally, the operation induces the constraints $\mathbf{s}_K \equiv \mathbf{s}_V$ on the pooling operators. In summary, \singleattnname is computed as, 
$$
\operatorname{PA}(\cdot) = \operatorname{Softmax}(\mathcal{P}(Q; \mathbf{\Theta}_Q)\mathcal{P}(K; \mathbf{\Theta}_K)^T/\sqrt{d})\mathcal{P}(V; \mathbf{\Theta}_V),
$$
where $\sqrt{d}$ is normalizing the inner product matrix row-wise. The output of the Pooling attention operation thus has its sequence length reduced by a \textit{stride} factor of $s^Q_{T}s^Q_{H}s^Q_{W}$ following the shortening of the query vector $Q$ in $\mathcal{P}(\cdot)$.

 \paragraph{Multiple heads.} As in  \cite{vaswani2017attention} the computation can be parallelized by considering $h$ heads where each head is performing the pooling attention on a non overlapping subset of $D/h$ channels of the $D$ dimensional input tensor $X$.
 
 \paragraph{Computational Analysis.}Since attention computation scales quadratically \wrt the sequence length, pooling the key, query and value tensors has dramatic benefits on the fundamental compute and memory requirements of the \fullnameCC model. Denoting the sequence length reduction factors by $f_Q$, $f_K$ and $f_V$ we have, 
 $$
f_j = s^j_T \cdot s^j_H \cdot s^j_W, \ \forall \ j \in \{Q,K,V\}.
$$
Considering the input tensor to $\mathcal{P}(; \Theta)$ to have dimensions $D \times T \times H \times W$, the run-time complexity of \attnabbv is $O(THW D/h (D+THW/f_Qf_K))$ per head and the memory complexity is $O(THWh(D/h + THW/f_Qf_K))$. 

This trade-off between the number of channels $D$ and sequence length term $THW/f_Qf_K$ informs our design choices about architectural parameters such as number of heads and width of layers. We refer the reader to the supplement for a detailed analysis and discussions on the time-memory complexity trade-off.

\subsection{\fullnameCC Networks}
\label{sec:pooling_transformers}
 Building upon \attnnameCC (\sref{sec:pooling_attn}), we describe the \fullnameCC model for visual representation learning using exclusively \attnabbv and MLP layers. 
 First, we present a brief review of the Vision Transformer Model that informs our design.

\begin{table}[t!]
		\vspace{-2.5em}
	\scriptsize
	\centering
	\captionsetup[subfloat]{captionskip=2pt}
	\captionsetup[subffloat]{justification=centering}
	
	\tablestyle{2pt}{1.05}
	\tablestyle{1pt}{1.08}
	\begin{tabular}{c|c|c}
		stage & operators & output sizes \\
		\shline
		\multirow{1}{*}{data layer} &  \multirow{1}{*}{stride \tcolor{$\tau$}\x\xycolor{1}\x\xycolor{1}}  &  \outsizesRaw{\tcolor{$T$}}{\xycolor{$H$}}{\xycolor{$W$}}{1}   \\
		\hline
		
		\multirow{2}{*}{patch$_1$} & \multicolumn{1}{c|}{\tcolor{1}\x\xycolor{16}\x\xycolor{16}, \wcolor{D}} &    \outsizesRawD{\wcolor{$D$}}{\tcolor{$T$}}{\xycolor{$\frac{H}{16}$}}{\xycolor{$\frac{W}{16}$}}{2}    \\
		& stride \tcolor{1}\x\xycolor{16}\x\xycolor{16}   \\
		\hline
		\multirow{2}{*}{scale$_2$}  & \blockatt{\wcolor{$D$}}{{$4D$}}{\dcolor{$N$}} &  \outsizesRawD{\wcolor{$D$}}{\tcolor{$T$}}{\xycolor{$\frac{H}{16}$}}{\xycolor{$\frac{W}{16}$}}{2}  \\
		&  & \\
	\end{tabular}
	\vspace{5pt}
	\caption{\textbf{Vision Transformers (ViT)} base model starts from a data layer that samples visual input with rate {$\tau$}\x{1}\x{1} to $T$\x$H$\x$W$  resolution, where $T$ is the number of frames $H$ height and $W$ width. The first layer, patch$_1$ projects patches (of shape  \tcolor{$1$}\x\xycolor{$16$}\x\xycolor{$16$}) to form a sequence, processed by a stack of \dcolor{$N$} transformer blocks (stage$_2$) at uniform channel dimension (\wcolor{$D$}) and resolution (\tcolor{${T}$}\x\xycolor{$\frac{H}{16}$}\x\xycolor{$\frac{W}{16}$}). }
	\label{tab:arch_vit_base}
\end{table}

\paragraph{Preliminaries: Vision Transformer (ViT).}
The Vision Transformer (ViT) architecture \cite{dosovitskiy2020image} starts by dicing the input video of resolution $T$\x$H$\x$W$, where $T$ is the number of frames $H$ the height and $W$ the width, into non-overlapping patches of size 1\x16\x16 each, followed by point-wise application of linear layer on the flattened image patches to to project them into the latent dimension, $D$, of the transformer. This is \textit{equivalent} to a convolution with equal kernel size and stride of  1\x16\x16 and is shown as patch$_1$ stage in the model definition in \tblref{tab:arch_vit_base}.

Next, a positional embedding $\mathbf{E} \in \mathbb{R}^{L \times D}$ is added to each element of the projected sequence of length $L$ with  dimension $D$ to encode the positional information and break permutation invariance. A learnable class embedding is appended to the projected image patches. 

The resulting  sequence of  length of $L+1$ is then processed sequentially by a stack of $N$ transformer blocks, each one performing attention ($\operatorname{MHA}$~ \cite{vaswani2017attention}), multi-layer perceptron ($\operatorname{MLP}$) and layer normalization ($\operatorname{LN}$\cite{Ba2016}) operations. Considering $X$ to be the input of the block, the output of a single transformer block, $\operatorname{Block}(X)$ is computed by
\begin{align*}
	\begin{split}
		X_1 &= \operatorname{MHA}(\operatorname{LN}(X)) + X \\
		\operatorname{Block}(X) &= \operatorname{MLP}(\operatorname{LN}(X_1)) + X_1. \\     
	\end{split}
\end{align*}
The resulting sequence after $N$ consecutive blocks is layer-normalized and the class embedding is extracted and passed through a linear layer to predict the desired output (\eg class). By default, the hidden dimension of the MLP is 4$D$. We refer the reader to\cite{dosovitskiy2020image, vaswani2017attention} for details.

In context of the present paper, it is noteworthy that ViT maintains a constant channel capacity and spatial resolution throughout all the blocks (see \tblref{tab:arch_vit_base}).

\begin{table}[t!]
		\vspace{-2.5em}
	\centering
	\tablestyle{1pt}{1.08}
	\begin{tabular}{c|c|c}
		stages & operators & output sizes \\ 
		\shline
		\multirow{1}{*}{data layer} & \multirow{1}{*}{stride \tcolor{$\tau$}\x\xycolor{1}\x\xycolor{1}}  &  \outsizesRawD{\wcolor{$D$}}{\tcolor{$T$}}{\xycolor{$H$}}{\xycolor{$W$}}{1}  \\
		\hline
		\multirow{2}{*}{cube$_1$} & \multicolumn{1}{c|}{\tcolor{$c_T$}\x\xycolor{$c_H$}\x\xycolor{$c_W$}, \wcolor{$D$}} &    \outsizesRawD{\wcolor{$D$}}{\tcolor{$\frac{T}{s_T}$}}{\xycolor{$\frac{H}{4}$}}{\xycolor{$\frac{W}{4}$}}{2}    \\
		& stride \tcolor{$s_T$}\x\xycolor{4}\x\xycolor{4}    \\
		\hline
		\multirow{2}{*}{scale$_2$}  & \blockatta{\wcolor{$D$}}{\wcolor{$4D$}}{\dcolor{$N_2$}} & \outsizesRawD{\wcolor{$D$}}{\tcolor{$\frac{T}{s_T}$}}{\xycolor{$\frac{H}{4}$}}{\xycolor{$\frac{W}{4}$}}{2} \\
		&  & \\
		\hline
		\multirow{2}{*}{scale$_3$}  & \blockatta{\wcolor{$2D$}}{\wcolor{$8D$}}{\dcolor{$N_3$}} & \outsizesRawD{\wcolor{$2D$}}{\tcolor{$\frac{T}{s_T}$}}{\xycolor{$\frac{H}{8}$}}{\xycolor{$\frac{W}{8}$}}{2}  \\
		&  & \\
		\hline
		\multirow{2}{*}{scale$_4$}  & \blockatta{\wcolor{$4D$}}{\wcolor{$16D$}}{\dcolor{$N_4$}} &\outsizesRawD{\wcolor{$4D$}}{\tcolor{$\frac{T}{s_T}$}}{\xycolor{$\frac{H}{16}$}}{\xycolor{$\frac{W}{16}$}}{2}   \\
		&  & \\
		\hline
		\multirow{2}{*}{scale$_5$}  & \blockatta{\wcolor{$8D$}}{{\wcolor{$32D$}}}{\dcolor{$N_5$}} & \outsizesRawD{\wcolor{$8D$}}{\tcolor{$\frac{T}{s_T}$}}{\xycolor{$\frac{H}{32}$}}{\xycolor{$\frac{W}{32}$}}{2}   \\
		&  & \\
	\end{tabular}
	\vspace{5pt}
	\caption{\textbf{Multiscale Vision Transformers (MViT)} base model.  Layer cube$_1$, projects \textit{dense} space-time cubes (of shape  \tcolor{$c_t$}\x\xycolor{$c_y$}\x\xycolor{$c_w$}) to \wcolor{$D$} channels  to reduce spatio-temporal resolution to \tcolor{$\frac{T}{s_T}$}\x\xycolor{$\frac{H}{4}$}\x\xycolor{$\frac{W}{4}$}. The subsequent stages progressively down-sample this resolution (at beginning of a stage) with \eicolor{MHPA} while simultaneously increasing the channel dimension, in MLP layers, (at the end of a stage). Each stage consists of \dcolor{$N_*$} transformer blocks, denoted in [brackets]. 
\label{tab:arch_MViT_base}
	\vspace{-15pt}
	}
\end{table}

\begin{table*}[t]
	\vspace{-3.0em}
	\footnotesize
	\tiny 
	\centering
	\captionsetup[subfloat]{captionskip=2pt}
	\captionsetup[subffloat]{justification=centering}
	\subfloat[\textbf{ViT}-B  with \textbf{179.6}G FLOPs, \textbf{87.2}M param, \textbf{16.8}G memory, and \textbf{68.5}\% top-1 accuracy.
	\label{tab:arch_vitb}]{
		\tablestyle{1pt}{1.05}
		\tablestyle{1pt}{1.08}  \scriptsize 
		\begin{tabular}{c|c|c}
			stage & operators & output sizes  \\
			\shline
			\multirow{1}{*}{data} &  \multirow{1}{*}{stride \tcolor{8}\x1\x1}   &  \outsizesRaw{\tcolor{8}}{\xycolor{224}}{\xycolor{224}}{1}   \\
			%			&  &  \\
			\hline
			
			\multirow{2}{*}{patch$_1$} & \multicolumn{1}{c|}{1\x16\x16, {768}} &    \outsizesRawD{\wcolor{$768$}}{\tcolor{8}}{\xycolor{14}}{\xycolor{14}}{2}    \\
			& stride 1\x16\x16   \\
			\hline
			\multirow{2}{*}{scale$_2$}  & \blockatt{768}{{3072}}{12} & \outsizesRawD{\wcolor{$768$}}{\tcolor{8}}{\xycolor{14}}{\xycolor{14}}{2}  \\
			&  & \\
			\hline
			\multicolumn{3}{c}{} \\ % space holder
			\multicolumn{3}{c}{} \\ % space holder
			\multicolumn{3}{c}{} \\ % space holder
			\multicolumn{3}{c}{} \\ % space holder
			\multicolumn{3}{c}{} \\ % space holder
			\multicolumn{3}{c}{} \\ % space holder
		\end{tabular}
	}	
	\hspace{6pt}
	\subfloat[\textbf{MViT}-B  with \textbf{70.5}G FLOPs, \textbf{36.5}M param, \textbf{6.8}G memory, and \textbf{77.2}\% top-1 accuracy.	
	\label{tab:arch_m}]{
		\tablestyle{0.5pt}{1.05}
		\tablestyle{1pt}{1.08}  \scriptsize 
		\begin{tabular}{c|c|c}
			stage & operators & output sizes \\
			\shline
			\multirow{1}{*}{data} & \multirow{1}{*}{stride \tcolor{4}\x1\x1}   &  \outsizesRaw{\tcolor{16}}{\xycolor{224}}{\xycolor{224}}{1}   \\
			\hline
			
			\multirow{2}{*}{cube$_1$} & \multicolumn{1}{c|}{3\x7\x7, {96}} &    \outsizesRawD{\wcolor{$96$}}{\tcolor{8}}{\xycolor{56}}{\xycolor{56}}{2}    \\
			& stride 2\x4\x4   \\
			\hline
			\multirow{2}{*}{scale$_2$}  & \blockatta{96}{{384}}{1} & \outsizesRawD{\wcolor{$96$}}{\tcolor{8}}{\xycolor{56}}{\xycolor{56}}{2}  \\
			&  & \\
			\hline
			\multirow{2}{*}{scale$_3$}  & \blockatta{192}{{768}}{2} & \outsizesRawD{\wcolor{$192$}}{\tcolor{8}}{\xycolor{28}}{\xycolor{28}}{2}  \\
			&  & \\
			\hline
			\multirow{2}{*}{scale$_4$}  & \blockatta{384}{{1536}}{11} & \outsizesRawD{\wcolor{$384$}}{\tcolor{8}}{\xycolor{14}}{\xycolor{14}}{2}  \\
			&  & \\
			\hline
			\multirow{2}{*}{scale$_5$}  & \blockatta{768}{{3072}}{2} & \outsizesRawD{\wcolor{$768$}}{\tcolor{8}}{\xycolor{7}}{\xycolor{7}}{2}  \\
			&  & \\
			\hline
		\end{tabular}%}
	}		
	\hspace{6pt}
	\subfloat[\textbf{MViT}-S  with \textbf{32.9}G FLOPs, \textbf{26.1}M param, \textbf{4.3}G memory, and \textbf{74.3}\% top-1 accuracy.
	\label{tab:arch_s}]{
		\tablestyle{0.5pt}{1.05}
		\tablestyle{1pt}{1.08}  \scriptsize 
		\begin{tabular}{c|c|c}
			stage & operators & output sizes  \\
			\shline
			\multirow{1}{*}{data} & \multirow{1}{*}{stride \tcolor{4}\x1\x1}  &  \outsizesRaw{\tcolor{16}}{\xycolor{224}}{\xycolor{224}}{1}   \\
			%			&  &  \\
			\hline
			
			\multirow{2}{*}{cube$_1$} & \multicolumn{1}{c|}{3\x8\x8, {128}} &    \outsizesRawD{\wcolor{$128$}}{\tcolor{8}}{\xycolor{28}}{\xycolor{28}}{2}    \\
			& stride 2\x8\x8   \\
			\hline
			\multirow{2}{*}{scale$_2$}  & \blockatta{128}{{512}}{3} & \outsizesRawD{\wcolor{$128$}}{\tcolor{8}}{\xycolor{28}}{\xycolor{28}}{2}  \\
			&  & \\
			\hline
			\multirow{2}{*}{scale$_3$}  & \blockatta{256}{{1024}}{7} & \outsizesRawD{\wcolor{$256$}}{\tcolor{8}}{\xycolor{14}}{\xycolor{14}}{2}  \\
			&  & \\
			\hline
			\multirow{2}{*}{scale$_4$}  & \blockatta{512}{{2048}}{6} & \outsizesRawD{\wcolor{$512$}}{\tcolor{8}}{\xycolor{7}}{\xycolor{7}}{2}  \\
			&  & \\
			\hline
			\multicolumn{3}{c}{} \\ % space holder
			\multicolumn{3}{c}{} \\ % space holder
		\end{tabular}%}
	}	
	\vspace{.1em}
	\caption{Comparing \textbf{ViT}-B to two instantiations of \textbf{MViT} with varying complexity,  \textbf{MViT}-S  in \protect\subref{tab:arch_s} and \textbf{MViT}-B in \protect\subref{tab:arch_m}. \textbf{MViT}-S operates at a lower spatial resolution and lacks a first high-resolution stage. The top-1 accuracy corresponds to \textit{5-Center} view testing on K400. FLOPs correspond to a single inference clip, and memory is for a training batch of 4 clips. See Table~\ref{tab:arch_MViT_base} for the general \textbf{MViT}-B structure. 
	}
	\label{tab:arch_instances}
	\vspace{-15pt}
\end{table*}

\paragraph{Multiscale Vision Transformers (MViT).} \label{sec:stage_concept}

Our key concept is to progressively \textit{grow} the \textit{channel} resolution (\ie dimension), while simultaneously \textit{reducing} the \textit{spatiotemporal}  resolution  (\ie sequence length) throughout the network. By design, our MViT architecture has \emph{fine} spacetime (and \textit{coarse} channel) resolution in early layers that is up-/downsampled to a coarse spacetime (and \textit{fine} channel) resolution in late layers. MViT is shown in \tblref{tab:arch_MViT_base}. 

\paragraph{Scale stages.}
A \textit{scale stage} is defined as a set of $N$ transformer blocks that operate on the same \textit{scale} with identical  resolution across channels and space-time dimensions $D$\x$T$\x$H$\x$W$.  At the input (cube$_1$ in \tblref{tab:arch_MViT_base}), we project the patches (or cubes if they have a temporal extent) to a smaller channel dimension (\eg 8\x~smaller than a typical ViT model), but long sequence (\eg 4\x4 $=$ 16\x~denser than a typical ViT model; \textit{cf.}~\tblref{tab:arch_vit_base}).

At a stage \textit{transition} (\eg scale$_1$ to  scale$_2$ to   in \tblref{tab:arch_MViT_base}), the channel dimension of the processed sequence is up-sampled while the length of the sequence is down-sampled. This effectively reduces the spatio-temporal resolution of the underlying visual data while allowing the network to assimilate the processed information in more complex features.

\paragraph{Channel expansion.} \label{sec:channel_growth}
 When transitioning from one stage to the next, we expand the channel dimension by increasing the output of the final MLP layer in the previous stage by a factor that is relative to the resolution change introduced at the stage. Concretely, if we down-sample the space-time resolution by 4\x, we increase the channel dimension by 2\x. For example, scale$_3$ to  scale$_4$ changes resolution from  $2D$\x$\frac{T}{s_T}$\x$\frac{H}{8}$\x$\frac{T}{8}$ to $4D$\x$\frac{T}{s_T}$\x$\frac{H}{16}$\x$\frac{T}{16}$ in \tblref{tab:arch_MViT_base}.  This roughly preserves the computational complexity across stages, and is similar to ConvNet design principles \cite{Simonyan2014,He2016}. 

 \paragraph{Query pooling.} The pooling attention operation affords flexibility not only in the length of key and value vectors but also in the length of the query, and thereby output, sequence. Pooling the query vector $\mathcal{P}(Q; \mathbf{k}; \mathbf{p}; \mathbf{s})$ with a kernel $\mathbf{s} \equiv (s^Q_T,s^Q_H,s^Q_W)$ leads to sequence reduction by a factor of $s^Q_T \cdot s^Q_H \cdot s^Q_W$. Since, our intention is to decrease resolution at the beginning of a stage and then preserve this resolution throughout the stage, only the first pooling attention operator of each stage operates at non-degenerate query stride $\mathbf{s}^Q > 1$, with all other operators constrained to $\mathbf{s}^Q \equiv (1,1,1)$.   

 \paragraph{Key-Value pooling.} Unlike Query pooling, changing the sequence length of key $K$ and value $V$ tensors, does not change the output sequence length and, hence, the space-time resolution. However, they play a key role in overall computational requirements of the pooling attention operator. 
 
 We decouple the usage of $K,V$ and $Q$ pooling, with $Q$ pooling being used in the first layer of each stage and $K,V$ pooling being employed in all other layers.
Since the sequence length of key and value tensors need to be identical to allow attention weight calculation, the pooling stride used on  $K$ and value $V$ tensors needs to be identical. In our default setting, we constrain \textit{all} pooling parameters ($\mathbf{k}; \mathbf{p}; \mathbf{s}$) to be identical \ie $\Theta_K \equiv \Theta_V$ within a stage, but vary  $\mathbf{s}$  \textit{adaptively} \wrt to the scale across stages.   

 \paragraph{Skip connections.} Since the channel dimension and sequence length change inside a residual block, we pool the skip connection to adapt to the dimension mismatch between its two ends. \attnabbv handles this mismatch by adding the query pooling operator $\mathcal{P}(\cdot ; \mathbf{\Theta}_Q)$ to the residual path. As shown in \figref{fig:pooling_attn}, instead of directly adding the input $X$ of MHPA to the output, we add the pooled input $X$ to the output, thereby matching the resolution to  attended query $Q$. 
 
 For handling the channel dimension mismatch between stage changes, we employ an extra linear layer that operates on the layer-normalized output of our \attnabbv operation. Note that this differs from the other (resolution-preserving) skip-connections that operate on the un-normalized signal. 

\subsection{Network instantiation details}

 \tblref{tab:arch_instances} shows concrete instantiations of the base models for Vision Transformers~\cite{dosovitskiy2020image} and our Multiscale Vision Transformers.  
 ViT-Base~\cite{dosovitskiy2020image} (\tblref{tab:arch_m}) initially projects the input to patches of shape 1\x16\x16 with dimension $D=768$, followed by stacking $N=12$ transformer blocks. With an 8\x224\x224 input  the resolution is fixed to 768\x8\x14\x14 throughout \textit{all} layers. The sequence length (spacetime resolution + class token) is $8 \cdot 14 \cdot 14 + 1 = 1569$.
 
 Our MViT-Base (\tblref{tab:arch_m}) is comprised of $4$ scale stages, each having several transformer blocks of consistent channel dimension. \textbf{MViT}-B initially projects the input to a channel dimension of $D=96$ with \textit{overlapping} space-time cubes of shape 3\x7\x7. The resulting sequence of length $8 * 56 *56 + 1 = 25089$  is reduced by a factor of $4$ for each additional stage, to a final sequence length of $8 * 7 *7 + 1 = 393$ at scale$_4$. In tandem, the channel dimension is up-sampled by a factor of $2$ at each stage, increasing to $768$ at scale$_4$. Note that all pooling operations, and hence the resolution down-sampling, is performed only on the data sequence without involving the processed class token embedding.

We set the number of \attnabbv heads to $h=1$ in the  scale$_1$ stage and increase the number of heads with the channel dimension (channels per-head  $D/h$ remain consistent at $96$).

At each stage transition, the previous stage output MLP dimension is increased by~2\x~and \attnabbv pools on $Q$ tensors with $\mathbf{s}^Q = (1,2,2)$ at the input of the next stage. 

We employ $K,V$ pooling in all \attnabbv blocks, with $\Theta_K \equiv \Theta_V$ and $\mathbf{s}^Q = (1,8,8)$ in scale$_1$ and \textit{adaptively} decay this stride \wrt to the scale across stages such that the $K,V$ tensors have consistent scale across all blocks.  

\section{Experiments: Video Recognition}

\paragraph{Datasets.} We use Kinetics-400 \cite{Kay2017} (K400) ($\app$240k training videos in 400 classes) and Kinetics-600~\cite{Carreira2017}. We further assess transfer learning performance for on   Something-Something-v2~\cite{ssv2}, Charades~\cite{Sigurdsson2016},  and AVA~\cite{Gu2018}.

We report top-1 and top-5 classification accuracy (\%) on the validation set, 
computational cost (in FLOPs) of a single, spatially center-cropped clip and the number of clips used. 

\paragraph{Training.} By default, all models are trained \emph{from random initialization} (``\emph{from scratch}'') on Kinetics, \emph{without} using ImageNet \cite{Deng2009} or other pre-training. Our training recipe and augmentations follow \cite{Feichtenhofer2019, touvron2020training}. For Kinetics, we train for 200 epochs with 2 repeated augmentation~\cite{hoffer2020augment} repetitions. 

We report  ViT baselines that are \textit{fine-tuned} from ImageNet, using a 30-epoch version of the training recipe in~\cite{Feichtenhofer2019}.

For the temporal domain, we sample a clip from the full-length video, and the input to the network are  $T$~frames with a temporal stride of $\tau$; denoted as $T\times\tau$~\cite{Feichtenhofer2019}.
\paragraph{Inference.}  We apply two testing strategies following~\cite{Feichtenhofer2019,feichtenhofer2020x3d}: \textit{(i)} Temporally, uniformly samples $K$ clips  (\eg $K$=$5$)  from a video, scales the shorter spatial side to 256~pixels and takes a 224\x224~center crop, and
\textit{(ii)}, the same as \textit{(i)} temporally, but take 3 crops of 224\x224~to cover the longer spatial axis.
We average the 
scores for all individual predictions.

All implementation specifics are in \S\ref{sec:app_training}.

\subsection{Main Results} \label{sec:results_main}
\begin{table}[t!]
		\vspace{-20pt}
	\hspace*{-7pt}
	\centering
	\tablestyle{2.0pt}{1.08}
		\resizebox{1.04\linewidth}{!}{
	\begin{tabular}{l|c|c|c|r|r}
		\multicolumn{1}{c|}{model}  &\multicolumn{1}{c|}{pre-train} &   top-1  & top-5  & {\scriptsize FLOPs\x views}  & Param \\
		\shline
		
		\hline		
		Two-Stream I3D \cite{Carreira2017}&   -&    71.6 & 90.0 & 216~\x~NA & 25.0 \\
		ip-CSN-152  \cite{Tran2019}  &  -&  77.8 & 92.8 & 109\x3\x10 & 32.8  \\
		{SlowFast} {\scriptsize 8\x 8 +NL}  \cite{Feichtenhofer2019}   &- & {78.7} & {93.5} &  116\x3\x10 & 59.9  \\ 
		{SlowFast} {\scriptsize{16\x 8 +NL}} \cite{Feichtenhofer2019}    &- &{79.8} & {93.9}  & 234\x3\x10 & 59.9  \\
		X3D-M~\cite{feichtenhofer2020x3d}  &- & 76.0  & 92.3 &  6.2\x3\x10  &  {3.8}  \\
		X3D-XL~\cite{feichtenhofer2020x3d}  &- & 79.1  & {93.9} &48.4\x3\x10 & 11.0  \\ 
		\shline
		{ViT-B-VTN \cite{neimark2021video}}  &  {ImageNet-1K} &    \demph{75.6} &  \demph{92.4} &  {4218\x1\x1} &  {114.0} \\
		{ViT-B-VTN \cite{neimark2021video}}  &  {ImageNet-\textbf{21K}} &    \demph{78.6} &  \demph{93.7} &  {4218\x1\x1} &  {114.0} \\
		{ViT-B-TimeSformer \cite{bertasius2021space}}  &  ImageNet-\textbf{21K} &  \demph{80.7} &  \demph{94.7} &  2380\x3\x1 &  {121.4} \\ 
		{ViT-L-ViViT \cite{arnab2021vivit}}  &  ImageNet-\textbf{21K} &  \demph{81.3} &  \demph{94.7} &  3992\x3\x4 & 310.8 \\ 
		\shline
		{ViT-B (our baseline)}  &  ImageNet-\textbf{21K}  &  \demph{79.3} & \demph{93.9} &  180\x1\x5   &  {87.2} \\ 
		{ViT}-B (our baseline) & - & 68.5 & 86.9 & 180\x1\x5   & 87.2 \\  \shline 
		\hline 
		\textbf{MViT}-S  & - & 76.0 & 92.1 &  32.9\x1\x5   & 26.1 \\ 
		
		\textbf{MViT}-B, 16\x4 & - & 78.4 & 93.5 & 70.5\x1\x5   & 36.6 \\ 
		
		\textbf{MViT}-B, 32\x3 & - & 80.2 & 94.4 & 170\x1\x5   & 36.6 \\ 
		\textbf{MViT}-B, 64\x3 & - & \textbf{81.2} &\textbf{95.1} & 455\x3\x3   & 36.6 \\ 
	\end{tabular}
		}
	\caption{\textbf{Comparison with previous work on Kinetics-400}. We report the inference cost with a single ``view" (temporal clip with spatial crop) $\times$ the number of views (FLOPs\x view$_\text{space}$\x view$_\text{time}$). Magnitudes are Giga ($10^9$) for FLOPs and Mega ($10^6$) for Param. Accuracy of models trained with external data is \demph{de-emphasized}.
	}
	\label{tab:sota:k400}
	\vspace{-1.4em}
\end{table}
%##################################################################################################

%##################################################################################################
\begin{table}[t!]
	\vspace{-20pt}
	\centering
	\small
	\tablestyle{2pt}{1.05}
	\begin{tabular}{l|c|c|c|r|r}
		\multicolumn{1}{c|}{model}  &\multicolumn{1}{c|}{pretrain} &   top-1  & top-5  & GFLOPs\x views  & Param \\
		\shline
		{SlowFast} {\scriptsize{16\x 8 +NL}} \cite{Feichtenhofer2019}  & - & {81.8}  & {95.1} &  234\x3\x10 & 59.9  \\
		{X3D-M}     &- & 78.8  & 94.5 &  6.2\x3\x10  & 3.8  \\
		{X3D-XL} &   -  & {81.9}  & {95.5} & 48.4\x3\x10 & 11.0  \\
		ViT-B-TimeSformer~\cite{bertasius2021space}  &  IN-\textbf{21K} & \demph{82.4} & \demph{96.0} &  1703\x3\x1 &  121.4  \\
		ViT-L-ViViT \cite{arnab2021vivit}  &  IN-\textbf{21K} &  \demph{83.0} &  \demph{95.7} &  3992\x3\x4 & 310.8 \\ 
		\hline 
		\textbf{MViT}-B, 16\x4 & - & 82.1 & 95.7 & 70.5\x1\x5   & 36.8 \\ 
		\textbf{MViT}-B, 32\x3 & - & {83.4} & \textbf{96.3} & 170\x1\x5  & 36.8 \\
		\textbf{MViT}-B-24, 32\x3 & - & \textbf{83.8} & \textbf{96.3} & 236\x1\x5  & 52.9 \\
	\end{tabular}
	%}
	\caption{\textbf{Comparison with previous work on Kinetics-600}. 
	}
	\label{tab:sota:k600}
	\vspace{-1.9em}
\end{table}
%##################################################################################################

\paragraph{Kinetics-400.} \tblref{tab:sota:k400} compares to prior work. 
From top-to-bottom, it has four sections and we discuss them in turn. 

The first \tblref{tab:sota:k400} section shows prior art using ConvNets. 

The second section shows concurrent work using Vision Transformers~\cite{dosovitskiy2020image} for video classification~\cite{neimark2021video,bertasius2021space}. Both approaches rely on ImageNet pre-trained base models. ViT-B-VTN \cite{neimark2021video} achieves 75.6\% top-1 accuracy, which is boosted by 3\% to 78.6\% by merely changing the pre-training from ImageNet-1K to ImageNet-21K. ViT-B-TimeSformer~\cite{bertasius2021space} shows another 2.1\% gain on top of VTN, at higher cost of 7140G FLOPs and 121.4M parameters. ViViT improves accuracy further with an even larger ViT-L model.

The third section in  \tblref{tab:sota:k400} shows our ViT baselines. We first list our ViT-B, also pre-trained on the ImageNet-21K, which achieves 79.3\%, thereby being 1.4\% lower than ViT-B-TimeSformer, but is with 4.4\x~fewer FLOPs and 1.4\x~fewer parameters. This result shows that \textit{simply fine-tuning an off-the-shelf ViT-B model from ImageNet-\textbf{21K}}~\cite{dosovitskiy2020image} provides a strong baseline on Kinetics. However, training this model from-scratch with the same fine-tuning recipe will result in 34.3\%. Using our ``training-from-scratch'' recipe will produce 68.5\% for this ViT-B model, using the  same 1\x5, spatial \x~temporal, views for video-level inference. 

The final section of \tblref{tab:sota:k400} lists our \textbf{MViT} results. All our models are \textit{trained-from-scratch} using this recipe, \textit{without} any external pre-training. Our small model, \mbox{\textbf{MViT}-S} produces 76.0\% while being relatively lightweight with  26.1M param and 32.9\x5${=}$164.5G FLOPs, outperforming ViT-B by \textbf{+7.5}\% at \textbf{5.5}\x~less compute in \textit{identical} train/val setting. 

Our base model, \textbf{MViT}-B provides 78.4\%, a \textbf{+9.9}\% accuracy boost over ViT-B under \textit{identical settings}, while having 2.6\x$/$2.4\x fewer FLOPs$/$parameters. When changing the frame sampling from 16\x4 to  32\x3 performance increases to 80.2\%. Finally, we take this model and fine-tune it for 5 epochs with longer 64 frame input, after interpolating the temporal positional embedding, to reach \textbf{81.2}\% top-1 using 3 spatial and 3 temporal views for inference (it is sufficient test with fewer temporal views if a clip has more frames).
Further quantitative and qualitative results are in \S\ref{sec:resultsapp}.

\paragraph{Kinetics-600}~\cite{Carreira2017} is a larger version of Kinetics. Results are in \tblref{tab:sota:k600}. We train MViT from-scratch, without any pre-training. \textbf{MViT}-B, 16\x4 achieves 82.1\% top-1 accuracy. We further train a deeper 24-layer model with longer sampling, \textbf{MViT}-B-24, 32\x3, to investigate model scale on this larger dataset. 
MViT achieves state-of-the-art of 83.4\% with 5-clip center crop testing while having 56.0\x~fewer FLOPs and 8.4\x~fewer parameters than ViT-L-ViViT \cite{arnab2021vivit} which relies on large-scale ImageNet-21K pre-training.

%%%% SSv2
\begin{table}[t] 
	\vspace{-15pt}
	\centering
	\small
	\tablestyle{2pt}{1.05}
	\resizebox{1\linewidth}{!}{
		\begin{tabular}{l|c|c|c|r|r}
			\multicolumn{1}{c|}{model}  &\multicolumn{1}{c|}{pretrain} &   top-1  & top-5  & FLOPs\x views  & Param \\
			\shline
			TSM-RGB~\cite{lin2019tsm}    & \scriptsize{IN-1K+K400} & 63.3 & 88.2 & 62.4\x3\x2 & 42.9\\  % from tsm codebase
			MSNet~\cite{kwon2020motionsqueeze} & IN-1K & 64.7 & 89.4 & 67\x1\x1 & 24.6 \\
			TEA~\cite{li2020tea}    & IN-1K & 65.1 & 89.9 & 70\x3\x10 & -\\
			{ViT-B-TimeSformer~\cite{bertasius2021space}}  &  IN-\textbf{21K} &  62.5 &  - &  1703\x3\x1 &  {121.4}  \\
			\shline
			{ViT-B (our baseline)}  &  IN-\textbf{21K}  &  63.5 & 88.3 &  180\x3\x1   &  {87.2} \\ \hline 
			
			SlowFast R50, 8\x8~\cite{Feichtenhofer2019} & \multirow{5}{*}{\scriptsize K400} & 61.9 & 87.0 &  65.7\x3\x1 & 34.1 \\
			SlowFast R101, 8\x8~\cite{Feichtenhofer2019} & & 63.1 & 87.6 & 106\x3\x1 & 53.3  \\
			\cline{1-1} \cline{3-6}
			\textbf{MViT}-B, 16\x4 & & 64.7	& 89.2 & 70.5\x3\x1 & 36.6 \\ 
			\textbf{MViT}-B, 32\x3 & & 67.1	& 90.8 & 170\x3\x1 & 36.6\\ 
			\textbf{MViT}-B, 64\x3 & & \textbf{67.7}	& \textbf{90.9} & 455\x3\x1 & 36.6\\ 
			\hline
			\textbf{MViT}-B, 16\x4 & \multirow{3}{*}{\scriptsize K600} & 66.2	& 90.2 & 70.5\x3\x1 & 36.6  \\ 
			\textbf{MViT}-B, 32\x3 & & 67.8	& 91.3 & 170\x3\x1 & 36.6\\ 
			\textbf{MViT}-B-24, 32\x3 & & \textbf{68.7}	& \textbf{91.5} & 236\x3\x1 & 53.2\\ 
		\end{tabular}
		% }
	}
	\caption{\textbf{Comparison with previous work on SSv2}.~~~~~~~~~~~~}

	\vspace{-1.4em}
	\label{tab:sota:ssv2}
\end{table} 

\paragraph{Something-Something-v2} (SSv2)~\cite{ssv2} is a dataset with videos containing object interactions, which is known as a `temporal modeling` task.  \tblref{tab:sota:ssv2} compares our method with the state-of-the-art. We first report a simple ViT-B (our baseline) that uses ImageNet-21K pre-training. 
Our \textbf{MViT}-B with 16 frames has 64.7\% top-1 accuracy, which is better than the SlowFast R101~\cite{Feichtenhofer2019} which shares the same setting (K400 pre-training and 3\x1 view testing). With more input frames, our \textbf{MViT}-B achieves 67.7\% and the deeper \mbox{\textbf{MViT}-B-24} achieves 68.7\% using our K600 pre-trained model of above. %the new state-of-the-art results with XX mAP 
In general, \tblref{tab:sota:ssv2} verifies the capability of temporal modeling for MViT.

%%%% Charades
\begin{table}[h] 
	\centering
	\small
	\tablestyle{2pt}{1.05}
	\begin{tabular}{l|x{40}|x{22}|r|r}
		\multicolumn{1}{c|}{model} &  \multicolumn{1}{c|}{pretrain} &  mAP    & \scriptsize FLOPs\x views & Param \\   
		\shline
		Nonlocal \cite{Wang2018} &  \multirow{2}{*}{\scriptsize IN-1K+K400}  &   37.5 & 544\x3\x10 & 54.3  \\
		STRG +NL \cite{Wang2018b} &   &   39.7 & 630\x3\x10 & 58.3 \\
		\hline
		Timeception \cite{hussein2019timeception} & \multirow{8}{*}{\scriptsize K400} & 41.1 &  N/A\x N/A & N/A \\
		LFB +NL \cite{Wu2019} &   & 42.5 & 529\x3\x10 & 122    \\
		{SlowFast} {\scriptsize{50, 8\x8}}~\cite{Feichtenhofer2019}  &   & 38.0  &  65.7\x3\x10  & 34.0 \\
		{SlowFast} {\scriptsize{101+NL, 16\x8}}~\cite{Feichtenhofer2019}  &  & 42.5  &  234\x3\x10  & 59.9  \\
		X3D-XL~\cite{feichtenhofer2020x3d} &     & 43.4 &  48.4\x3\x10 & 11.0   \\ 		
		\cline{1-1} \cline{3-5}
		\textbf{MViT}-B, 16\x4 &  & 40.0 & 70.5\x3\x10 & 36.4 \\ 
		\textbf{MViT}-B, 32\x3 &    &  44.3 & 170\x3\x10 & 36.4 \\ 
		\textbf{MViT}-B, 64\x3 &    &  \textbf{46.3} & 455\x3\x10 & 36.4 \\ 
		\shline
		{SlowFast} {\scriptsize{R101+NL, 16\x8}}~\cite{Feichtenhofer2019} &  \multirow{5}{*}{\scriptsize K600} &  {45.2} &  234\x3\x10  & 59.9  \\ 
		X3D-XL~\cite{feichtenhofer2020x3d} &    & 47.1 &  {48.4\x3\x10 }& {11.0}  \\ 
		\cline{1-1} \cline{3-5}
		\textbf{MViT}-B, 16\x4 &  & 43.9 & 70.5\x3\x10 & 36.4 \\ 
		\textbf{MViT}-B, 32\x3 &    &  47.1 & 170\x3\x10 & 36.4 \\ 
		\textbf{MViT}-B-24, 32\x3 &    &  \textbf{47.7} & 236\x3\x10 & 53.0 \\ 
	\end{tabular}
	
	\caption{\textbf{Comparison with previous work on Charades}. ~~~~~~~~~~~~~~~~~
	} 
	\vspace{-.9em}
	\label{tab:sota:charades}
\end{table} 

%##################################################################################################
\paragraph{Charades} \cite{Sigurdsson2016} is a dataset with longer range activities. We validate our model in \tblref{tab:sota:charades}. With similar FLOPs and parameters, our \textbf{MViT}-B 16\x4 achieves better results (+2.0 mAP) than SlowFast R50~\cite{Feichtenhofer2019}. As shown in the Table, the performance of \textbf{MViT}-B is further improved by increasing the number of input frames and \textbf{MViT}-B layers and using K600 pre-trained models.

%%%% AVA
\begin{table}[h!]
	\vspace{-15pt}
	\centering
	\tablestyle{2.5pt}{1.05}
	\begin{tabular}{l|c|c|r|r}
		\multicolumn{1}{c|}{model} & \multicolumn{1}{c|}{pretrain} &   val mAP  & FLOPs & Param \\ 
		\shline
		{SlowFast}, 4\x16, R50 \cite{Feichtenhofer2019} & \multirow{5}{*}{K400} & 21.9 & 52.6 & 33.7 \\ 
		SlowFast, 8\x8, R101~\cite{Feichtenhofer2019}   &  & 23.8  & 138 & 53.0 \\ 
		\cline{1-1} \cline{3-5}
		\textbf{MViT}-B, 16\x4   &  & 24.5 & 70.5 & 36.4 \\ 
		\textbf{MViT}-B, 32\x3    &   & 26.8 & 170 & 36.4 \\  
		\textbf{MViT}-B, 64\x3    &   & \textbf{27.3} & 455 & 36.4 \\  
		\shline
		
		{SlowFast}, 8\x8 R101+NL \cite{Feichtenhofer2019}   & \multirow{6}{*}{K600}   &  27.1 & 147 & 59.2 \\ 
		{SlowFast}, 16\x8 R101+NL \cite{Feichtenhofer2019}   &    &  27.5 & 296 & 59.2 \\
		X3D-XL~\cite{feichtenhofer2020x3d}  &    &  27.4 & 48.4 & 11.0 \\
		\cline{1-1} \cline{3-5}
		\textbf{MViT}-B, 16\x4   &  & 26.1 & 70.5 & 36.3 \\ 
		\textbf{MViT}-B, 32\x3    &   & 27.5 & 170 & 36.4 \\ 
		\textbf{MViT}-B-24, 32\x3    &   & \textbf{28.7} & 236& 52.9 \\  
		
	\end{tabular}
	
	\caption{\textbf{Comparison with previvous work on AVA v2.2}. All methods use \textit{single center crop} inference following~\cite{feichtenhofer2020x3d}.
	}
	\label{suptab:sota:ava}
	\vspace{-.5em}
\end{table}

\paragraph{AVA} \cite{Gu2018} is a dataset with for spatiotemporal-localization of human actions. We validate our MViT on this detection task. Details about the detection architecture of MViT can be found in \S\ref{sec:detection}.
Table~\ref{suptab:sota:ava} shows the results of our MViT models compared with SlowFast~\cite{Feichtenhofer2019} and X3D~\cite{feichtenhofer2020x3d}.
We observe that MViT-B can be competitive to SlowFast and X3D using the same pre-training and testing strategy. 

\subsection{Ablations on Kinetics}\label{sec:kinetics_details}
We carry out ablations on Kinetics-400 (K400) using 5-clip center 224\x224 crop testing. 
  We show top-1 accuracy (Acc), as well as computational complexity measured in GFLOPs
  for a  single clip input of spatial size 224$^2$. Inference computational cost is proportional as a fixed number of 5 clips is used (to roughly cover the inferred videos with $T$\x$\tau$=16\x4 sampling.) We also report  Parameters in M($10^6$) and training GPU memory in G($10^9$) for a  batch size of 4. By default all  MViT ablations are with \textbf{MViT}-B, $T$\x$\tau{=}$16\x4 and max-pooling in MHSA.

\begin{table}[h!]
	\centering
	\small
	\tablestyle{8pt}{1.1}
	\begin{tabular}{r|x{22}x{36}x{34}|l}
		model & shuffling   & FLOPs (G) & Param (M) & Acc  \\
		\shline
		\textbf{MViT}-B  & & \multirow{2}{*}{70.5} &  \multirow{2}{*}{36.5} & \textbf{77.2} \\
		\textbf{MViT}-B  & \checkmark  &  &  & {70.1}\macc{$-$7.1} \\ \hline 
		ViT-B &  & \multirow{2}{*}{179.6} &  \multirow{2}{*}{87.2}  & 68.5 \\ 
		ViT-B   & \checkmark  &  && 68.4\macc{$-$0.1} \\ 
	\end{tabular}
	\caption{\textbf{Shuffling frames in inference.} \textbf{MViT}-B severely drops ($-$7.1\%) for shuffled temporal input, but ViT-B models appear to \textit{ignore} temporal information as accuracy remains similar ($-$0.1\%).
		\label{tab:shuffling}
	}\vspace{-20pt}
\end{table}

\paragraph{Frame shuffling.}
Table~\ref{tab:shuffling} shows results for randomly shuffling the input frames in time during testing. All models are trained without any shuffling and have temporal embeddings. We notice that our \textbf{MViT}-B architecture suffers a significant accuracy drop of \textbf{-7.1}\% (77.2 $\rightarrow$ 70.1) for shuffling inference frames. By contrast, ViT-B is surprisingly robust for shuffling the temporal order of the input.

This indicates that a na\"ive application of ViT to video does not model temporal information, and the temporal positional embedding in ViT-B seems to be fully ignored. We also verified this with the 79.3\% \mbox{ImageNet-21K} pre-trained ViT-B of \tblref{tab:sota:k400}, which has \textit{\textbf{ the same accuracy}} of 79.3\% for shuffling test frames, suggesting that it implicitly performs bag-of-frames video classification in Kinetics.  

\begin{table}[h!]
	\vspace{-5pt}
	\centering
	\small
	\tablestyle{2pt}{1.1}
	\begin{tabular}{l|x{28}|ll|l}
		variant &{\scriptsize[{N$_1$}, {N$_2$}]}  & FLOPs (G) & Mem (G) & Acc \\
		\shline
		ViT-B & [12, 0] & 179.6  & 16.8 & 68.5 \\ \hline 
		2-scale ViT-B, $Q$ pool & [6, 6] &  \textbf{111.1}\pacc{$-$68.5} & 9.8\pacc{$-$7.0} & \textbf{71.0}\pacc{$+$1.5} \\ 
		ViT-B, $K,V$ pool & [12, 0] & 148.4\pacc{$-$31.2}  & \textbf{8.9}\pacc{$-$7.9}  & 69.1\pacc{$+$0.6}  \\ 
	\end{tabular}
	\caption{\textbf{Query (scale stage) and Key-Value pooling on ViT-B.} Introducing a \textit{single} extra resolution stage into ViT-B boosts accuracy by +1.5\%.  Pooling $K, V$ provides +0.6\% accuracy. Both techniques allow dramatic FLOPs/memory savings. 
	\label{tab:ablation:vit_pooling}
	} 
	\vspace{-5pt}
\end{table}

\paragraph{Two scales in ViT.}
We provide a simple experiment that ablates the effectiveness of scale-stage design on ViT-B. For this we add a \textit{single scale stage} to the ViT-B model. To isolate the effect of having different scales in ViT, we do not alter the channel dimensionality for this experiment. We do so by performing $Q$-Pooling with $\mathbf{s}^Q \equiv (1,2,2)$ after 6 Transformer blocks (\textit{cf.}~\tblref{tab:arch_instances}).
\tblref{tab:ablation:vit_pooling} shows the results. Adding a single scale stage to the ViT-B baseline boosts accuracy by +1.5\%  while deceasing FLOPs and memory cost by  38\% and 41\%. Pooling Key-Value tensors reduces compute and memory cost while slightly increasing accuracy.

\begin{table}[h!]
	\vspace{-5pt}
	\centering
	\small
	\tablestyle{8pt}{1.1}
	\begin{tabular}{lr|x{40}|x{22}}
		&positional embedding   & Param (M)  & Acc  \\
		\shline
		(i) &none   & 36.2 & 75.8 \\ 
		(ii) &space-only & 36.5 & 76.7  \\
		(iii)& joint space-time & 38.6 & 76.5  \\ 
		(iv)&\textbf{separate} in space \& time & 36.5 & \textbf{77.2} \\
	\end{tabular}
	\vspace{.5em}
	\caption{\textbf{Effect of separate space-time positional embedding.} Backbone: \textbf{MViT}-B, 16\x4. FLOPs are 70.5G for all variants.  
		\label{tab:pos_embed}
	}
	\vspace{-10pt}
\end{table}

\paragraph{Separate space \& time embeddings in MViT.} In \tblref{tab:pos_embed}, we ablate using (i) none, (ii) space-only, (iii) joint space-time, and (iv) a separate space and time (our default), positional embeddings. We observe that no embedding (i) decays accuracy by -0.9\% over using just a spatial one (ii) which is roughly equivalent to a joint spatiotemporal one (iii). Our separate space-time embedding (iv) is best, and also has 2.1M fewer parameters than a joint spacetime embedding. 

\begin{table}[h]
	\centering
	\small
	\tablestyle{8pt}{1.1}
	\tablestyle{2pt}{1.05}
	\begin{tabular}{r|x{50}|x{50}|x{24}x{24}|x{24}}
		\multicolumn{1}{c|}{ $T\times\tau$ } & $c_T{\times} c_H{\times} c_W$ & $s_T{\times} s_H{\times} s_W$ & FLOPs &Param  & Acc  \\
		\shline
		8\x 8 & 1\x4\x4 & 1\x4\x4 & 69.4 & 36.5 & 74.5 \\ 
		8\x 8 & 1\x7\x7 & 1\x4\x4 & 69.6 & 36.5 & 75.6 \\ 
		8\x 8 & 3\x7\x7 & 1\x4\x4 & 70.5 & 36.5  & 75.9 \\ 
		\underline{16\x4} & \underline{3\x7\x7} & \underline{2\x4\x4} & 70.5 & 36.5  & 77.2 \\
		32\x 2 & 3\x7\x7 & 4\x4\x4 & 70.5 & 36.5 & 77.2 \\ 
		32\x 2 & 7\x7\x7 & 4\x4\x4 & 70.5 & 36.5 & 77.3 \\ 
	\end{tabular}
	\vspace{.5em}
	\caption{\textbf{Input sampling}: We vary sampling rate $T\times\tau$, the size $\mathbf{c}{=}c_T{\times} c_H{\times} c_W$ and stride of $\mathbf{s}{=}s_T{\times} s_H{\times} s_W$ the cube$_1$ layer that projects space-time cubes. Cubes with temporal extent  $c_T>1$  are  beneficial.  Our default setting is \underline{underlined}. }
	\label{tab:sampling_rate}
		\vspace{-10pt}
\end{table}

\paragraph{Input Sampling Rate.} Table \ref{tab:sampling_rate} shows results for different cubification kernel size $\mathbf{c}$ and sampling stride $\mathbf{s}$ (\textit{cf.}~\tblref{tab:arch_MViT_base}). We observe that sampling \textit{patches}, $\mathbf{c}_T$ = 1, performs worse than sampling \textit{cubes} with $\mathbf{c}_T >$  1. Further, sampling twice as many frames, $T {=}$ 16, with twice the cube stride, $s_T {=}$ 2, keeps the cost constant but boosts performance by +1.3\% (75.9\% $\rightarrow$ 77.2\%). Also, sampling \textit{overlapping} input cubes $\mathbf{s} < \mathbf{c}$ allows better information flow and benefits performance. While $\mathbf{c}_T >$ 1 helps, very large temporal kernel size ($\mathbf{c}_T =$  7) doesn't futher improve performance.

\begin{table}[h]
	\centering
	\small
		\tablestyle{3pt}{1.05}
		\begin{tabular}{c|x{55}|x{22}x{22}x{22}|x{22}}
			variant &{\scriptsize[{N$_2$}, {N$_3$}, {N$_4$}, {N$_5$}]} & FLOPs & Param & Mem & Acc \\
			\shline
			V1 & [2, 6, 6, 2] & 90.2 & \textbf{29.5} & 11.0 & 76.3  \\ 
			V2 & [2, 4, 6, 4] & 86.9 & 42.8 & 10.3 & 75.9 \\ 
			V3 & [2, 4, 8, 2] & 88.3 & 32.2 & 10.5 & 76.6 \\ 
			V4 & [2, 2, 8, 4] & 85.0 & 45.5 & 9.7 &  76.7 \\ 
			\underline{V5} & \underline{[1, 2, 11, 2]} & \textbf{83.6} & 36.5 & \textbf{9.1} &\textbf{77.1} \\ 
			V6 & [2, 2, 10, 2] & 86.4 & 34.9 & 11.3 & 76.9 \\ 
	\end{tabular}
	\vspace{.5em}
	\caption{\textbf{Scale blocks}: We ablate the stage configuration as the number of blocks $N$ in stages of \textbf{MViT}-B (\ie where to pool $Q$). The overall number of  transformer blocks is constant with $N{=}16$.
	\label{tab:ablation:q_pool} }
		\vspace{-10pt}
\end{table}

\paragraph{Stage distribution.} The ablation in \tblref{tab:ablation:q_pool} shows the results for distributing the number of transformer blocks in each  individual scale stage. The overall number of transformer blocks, $N{=}16$ is consistent.  We observe that having more blocks in early stages increases memory and having more blocks later stages the parameters of the architecture. Shifting the majority of blocks to the scale$_4$ stage (Variant V5 and V6 in \tblref{tab:ablation:q_pool}) achieves the best trade-off. 

\begin{table}[h]
	\centering
	\small
	\vspace{5pt}
		\tablestyle{3pt}{1.05}
		\begin{tabular}{c|x{26}|x{22}x{22}|x{22}}
			\multicolumn{1}{c|}{stride $\mathbf{s}$ } & adaptive & FLOPs  & Mem & Acc \\
			\shline
			none &  n/a  & 130.8 & 16.3 & \textbf{77.6} \\ 
			1\x4\x4 & & 71.4 & 8.2  & 75.9 \\ 
			2\x4\x4 & & 64.3 & 6.6  & 74.8 \\ 
			2\x4\x4 & \checkmark & 83.6 & 9.1 & {77.1} \\ 
			\underline{1\x8\x8} & \underline{\checkmark}  & 70.5 & 6.8 & {77.2} \\ 
			2\x8\x8 & \checkmark & \textbf{63.7} &\textbf{6.3} & 75.8 \\ 
	\end{tabular}
	\vspace{.5em}
	\caption{\textbf{Key-Value pooling}: Vary stride  $\mathbf{s}=s_T \times s_H \times s_W$, for pooling $K$ and $V$. ``adaptive'' reduces stride \wrt stage resolution. \label{tab:ablation:kv_pool} }
		\vspace{-10pt}
\end{table}

\paragraph{Key-Value pooling.} The ablation in \tblref{tab:ablation:kv_pool} analyzes the pooling stride  $\mathbf{s}=s_T \times s_H \times s_W$, for pooling $K$ and $V$ tensors. Here, we compare an ``adaptive'' pooling that uses a stride \wrt stage resolution, and keeps the $K,V$ resolution \textit{fixed} across all stages, against a non-adaptive version that uses the same stride at every block. First, we compare the baseline which uses no $K,V$ pooling with non-adaptive pooling with a fixed stride of 2\x4\x4 across all stages: this drops accuracy from 77.6\% to 74.8 (and reduces FLOPs and memory by over 50\%). Using an adaptive stride that is 1\x8\x8 in the scale$_1$ stage, 1\x4\x4 in scale$_2$, and  1\x2\x2 in scale$_3$ gives the best accuracy of 77.2\% while still preserving most of the efficiency gains in FLOPs and memory.  
\begin{table}[h]
	\centering
	\small
		\tablestyle{3pt}{1.05}
		\tablestyle{3pt}{1.05}
		\begin{tabular}{c|x{48}|x{26}|x{26}}
			\multicolumn{1}{c|}{  kernel $\mathbf{k}$} & pooling func & Param & Acc  \\
			\shline
			$\mathbf{s}$ & max & 36.5 &{76.1} \\ 
			$2\mathbf{s}+1$ & max & 36.5 & {75.5} \\ 
			{$\mathbf{s}+1$}& {max}   & 36.5 & {77.2} \\
			$\mathbf{s}+1$ & average   & 36.5 &  75.4 \\ 
			$\mathbf{s}+1$ & conv & 36.6  &  78.3 \\ 
			3\x3\x3 & conv  & 36.6  &  \textbf{78.4}  \\ 
    	\end{tabular}
	\vspace{.5em}
	\caption{\textbf{Pooling function}: Varying the kernel $\mathbf{k}$ as a function of stride $\mathbf{s}$. Functions are average or max pooling and conv which is a learnable, channel-wise convolution.  \label{tab:ablation:pooling_k} }
		\vspace{-10pt}
\end{table}
\paragraph{Pooling function.}  The ablation in \tblref{tab:ablation:pooling_k} looks at the kernel size $\mathbf{k}$ \wrt the stride $\mathbf{s}$, and the pooling function (max/average/conv). First, we see that having equivalent kernel and stride  $\mathbf{k}{=}\mathbf{s}$ provides 76.1\%, increasing the kernel size to $\mathbf{k}{=}2\mathbf{s}+1$ decays to 75.5\%, but using a kernel $\mathbf{k}{=}\mathbf{s}+1$ gives a clear benefit of 77.2\%. This indicates that \textit{overlapping pooling is effective}, but a too large overlap ($2\mathbf{s}+1$) hurts. Second, we investigate average instead of max-pooling and observe that accuracy decays by  from 77.2\% to 75.4\%. 

Third, we use conv-pooling by a learnable, channelwise convolution followed by LN. This variant has +1.2\% over max pooling and is used for all experiments in \S\ref{sec:results_main} and \S\ref{sec:imagenet}.

\begin{table}[h]
	\centering
	\small
	\tablestyle{6pt}{1.1}
	\begin{tabular}{l|r|x{22}|r|r}
		model   & clips/sec &  Acc & {\scriptsize FLOPs\x views}  & Param \\ 
		\shline
		X3D-M~\cite{feichtenhofer2020x3d}    & 7.9 & 74.1 & 4.7\x1\x5 & 3.8  \\  
		SlowFast R50~\cite{Feichtenhofer2019} & 5.2 & 75.7 & 65.7\x1\x5 & 34.6 \\  
		SlowFast R101~\cite{Feichtenhofer2019} & 3.2 & 77.6 & 125.9\x1\x5 & 62.8\\  
		{ViT}-B~\cite{dosovitskiy2020image} & 3.6 & 68.5 &  179.6\x1\x5 & 87.2 \\\hline   
		\textbf{MViT}-S, max-pool  & \textbf{12.3} & 74.3 &  32.9\x1\x5   & 26.1 \\
		\textbf{MViT}-B, max-pool & 6.3 & 77.2& 70.5\x1\x5   & 36.5 \\ 
		\textbf{MViT}-S, conv-pool  & 9.4 & 76.0 &  32.9\x1\x5   & 26.1 \\
		\textbf{MViT}-B, conv-pool & 4.8 & 78.4 & 70.5\x1\x5   & 36.6 \\ 
	\end{tabular} 
%	\vspace{.5em}
	\caption{\textbf{Speed-Accuracy tradeoff on Kinetics-400. } Training throughput is measured in clips/s. MViT is \textit{fast} and \textit{accurate}. 
		\label{suptab:speed_acc}
	}
	\vspace{-1.2em}
\end{table}
\paragraph{Speed-Accuracy tradeoff.} In Table \ref{suptab:speed_acc}, we analyze the speed/accuracy trade-off of our MViT models, along with their counterparts vision transformer (ViT~\cite{dosovitskiy2020image}) and  ConvNets (SlowFast 8\x8 R50, SlowFast 8\x8 R101~\cite{Feichtenhofer2019}, \& X3D-L~\cite{feichtenhofer2020x3d}). We measure training throughput as the number of video clips per second on a single M40 GPU. 

We observe that both MViT-S and MViT-B models are not only significantly more accurate but also much faster than both the ViT-B baseline and convolutional models. Concretely, MViT-S  has \textbf{3.4\x} higher throughput speed (clips/s), is \textbf{+5.8}\% more accurate (Acc), and has \textbf{3.3\x} fewer parameters (Param) than ViT-B. Using a conv instead of max-pooling in MHSA, we observe a training speed reduction of $\app$20\% for convolution and additional parameter updates. 

\section{Experiments: Image Recognition}\label{sec:imagenet}

We apply our video models on static image recognition by using them with single frame, $T=1$, on ImageNet-1K~\cite{Deng2009}.

\paragraph{Training.} Our recipe is identical to DeiT~\cite{touvron2020training} and summarized in the supplementary~material. Training  is for 300 epochs and results improve for training longer~\cite{touvron2020training}.

\subsection{Main Results}
For this experiment, we take our models which were designed by ablation studies for video classification on Kinetics and \textit{simply remove the temporal dimension}. Then we train and validate them (``from scratch'') on ImageNet.

Table \ref{tab:sota:in1k} shows the comparison with previous work. From top to bottom, the table contains RegNet \cite{ilija_2020}
and EfficientNet \cite{tan2019efficientnet} 
as ConvNet examples, and 
DeiT \cite{touvron2020training}, with  DeiT-B  being identical to ViT-B~\cite{dosovitskiy2020image} but trained with the improved recipe in~\cite{touvron2020training}. Therefore, this is the vision transformer counterpart we are interested in comparing to. 

The bottom section in \tblref{tab:sota:in1k} shows results for our Multiscale Vision Transformer (MViT) models. 

We show models of different depth, \textbf{MViT}-B-Depth, (16, 24, and 32), where \textbf{MViT}-B-16 is our base model and the deeper variants are simply created by repeating the number of blocks $N_*$ in each scale stage (\textit{cf.}~\tblref{tab:arch_m}). ``wide'' denotes a larger channel dimension of $D=112$. All our models are trained using the identical recipe as DeiT \cite{touvron2020training}.

We make the following observations: 

(i) Our lightweight \textbf{MViT}-B-16 achieves 82.5\% top-1 accuracy, with only 7.8 GFLOPs, which outperforms the DeiT-B counterpart by +0.7\% with lower computation cost (2.3\x fewer FLOPs and Parameters). If we use conv instead of max-pooling, this number is increased by +0.5\% to 83.0\%.

(ii) Our deeper model \textbf{MViT}-B-24, provides a gain of {+0.6}\% accuracy at slight increase in computation. 

(iii) A  larger model, \textbf{MViT}-B-24-wide with input resolution 320$^2$ reaches 84.3\%, corresponding to a +1.2\% gain, at 1.7\x fewer FLOPs, over DeiT-B$\uparrow$384$^2$. Using convolutional, instead of max-pooling elevates this to  \textbf{84.8}\%.

These results suggest that Multiscale Vision Transformers have an architectural advantage over Vision Transformers.  

\begin{table}[t!]
%	\vspace{-10pt}
	\centering
	\tablestyle{1.8pt}{1.05}
	\begin{tabular}{l|c|c|r}
		\multicolumn{1}{c|}{model}  &  Acc  & FLOPs (G)   & Param (M) \\
		\shline		
		\hline
		{RegNetZ-4GF \cite{dollar2021fast} }  &   83.1 & 4.0 &  28.1 \\ 
		{RegNetZ-16GF \cite{dollar2021fast} }  &   84.1 & 15.9 &  95.3 \\
		{EfficientNet-B7 \cite{tan2019efficientnet}}  &  84.3 & 37.0 & 66.0 \\
		\hline
		DeiT-S \cite{touvron2020training} &     79.8 & 4.6 & 22.1 \\
		DeiT-B \cite{touvron2020training} &     81.8 & 17.6 & 86.6 \\
		DeiT-B $\uparrow$ 384$^{2}$ \cite{touvron2020training} & 83.1 &  55.5 &   87.0 \\

		\shline

		\textbf{MViT}-B-16, max-pool &     82.5 & 7.8 & 37.0 \\
		\textbf{MViT}-B-24, max-pool &     83.1 & 10.9 & 53.5 \\
		\textbf{MViT}-B-24-wide-320$^{2}$, max-pool &  84.3  & 32.7 & 72.9 \\ 
		
		\hline	
		\textbf{MViT}-B-16 &     83.0 & 7.8 & 37.0 \\
		\textbf{MViT}-B-24-wide-320$^{2}$ &  \textbf{84.8}  &  32.7 & 72.9 \\ 
		
	\end{tabular}% }
	\vspace{3pt}
	\caption{\textbf{Comparison to prior work on ImageNet}. \mbox{RegNet} and EfficientNet are ConvNet examples that use different training recipes. DeiT$/$MViT are ViT-based and use identical recipes~\cite{touvron2020training}.  
	}
	\label{tab:sota:in1k}
	\vspace{-1.0em}
\end{table}

\section{Conclusion}
We have presented Multiscale Vision Transformers  that aim to connect the fundamental concept of multiscale feature hierarchies with the transformer model. MViT hierarchically expands the feature complexity while reducing visual resolution. In empirical evaluation, MViT shows a fundamental advantage over single-scale vision transformers for video and image recognition.   
We hope that our approach will foster further research in visual recognition.

\newcount\cvprrulercount
\appendix
\section*{Appendix}
\setcounter{table}{0}
\renewcommand{\thetable}{A.\arabic{table}}
\renewcommand{\thefigure}{A.\arabic{figure}}

In this appendix, 
\S\ref{sec:resultsapp} contains further 
\textit{ablations} for  Kinetics (\S\ref{sec:ablations_kinetics}) \& ImageNet (\S\ref{sec:ablations_imagenet}), 
\S\ref{sec:analysis} contains an \textit{analysis} on computational complexity of \attnabbvspace, and \S\ref{sec:ablations_qualitative} qualitative \textit{observations} in MViT and ViT models.  
\S\ref{sec:app_training} contains additional \textit{implementation details} for:  Kinetics (\S\ref{sec:kineticsapp}), AVA  (\S\ref{sec:detection}),  Charades (\S\ref{sec:charades}), SSv2 (\S\ref{sec:ssv2}), and ImageNet (\S\ref{sec:details_imagenet}).

\begin{figure*}[t]
	\centering
	\includegraphics[width=0.46\linewidth]{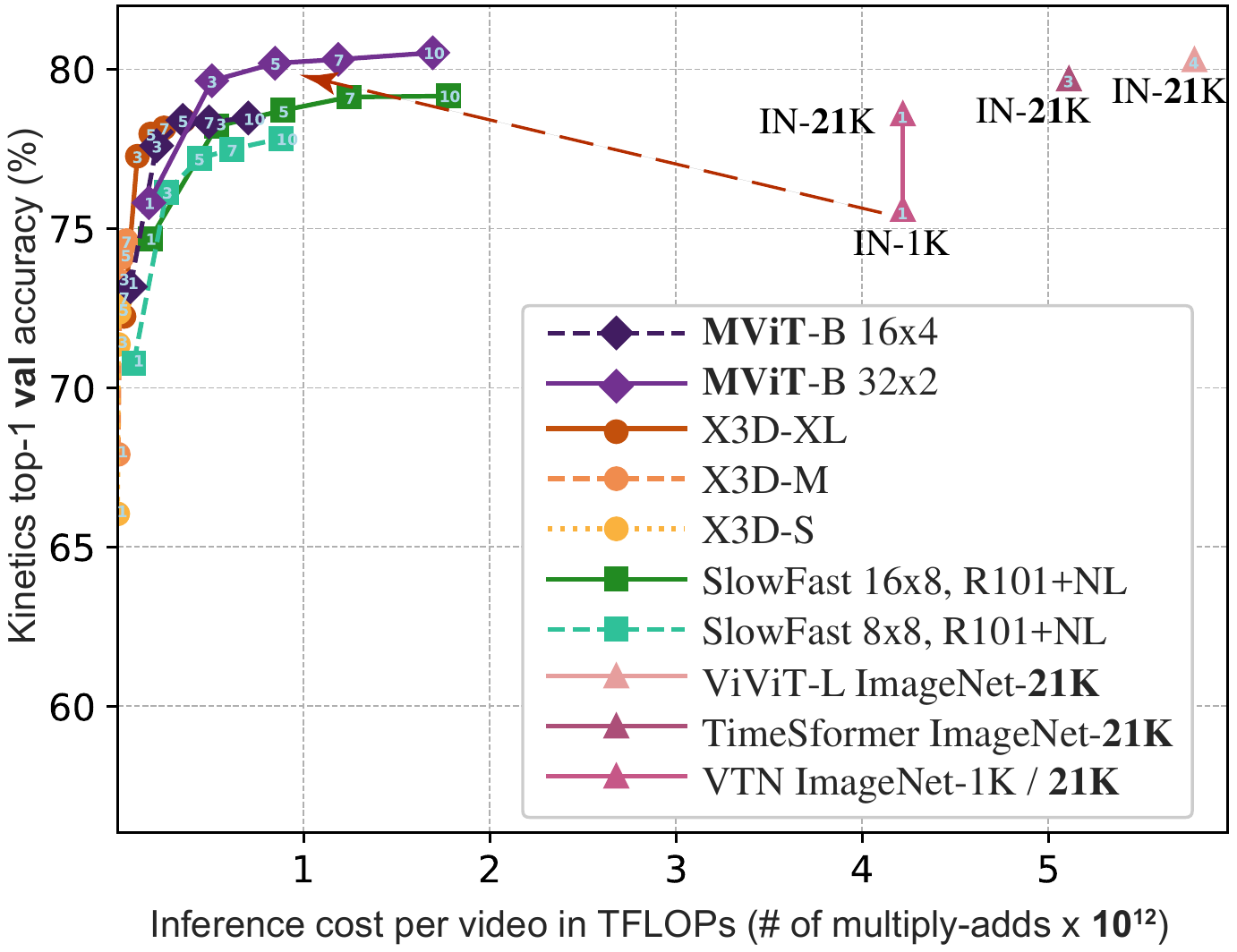} \hfill
	\includegraphics[width=0.47\linewidth]{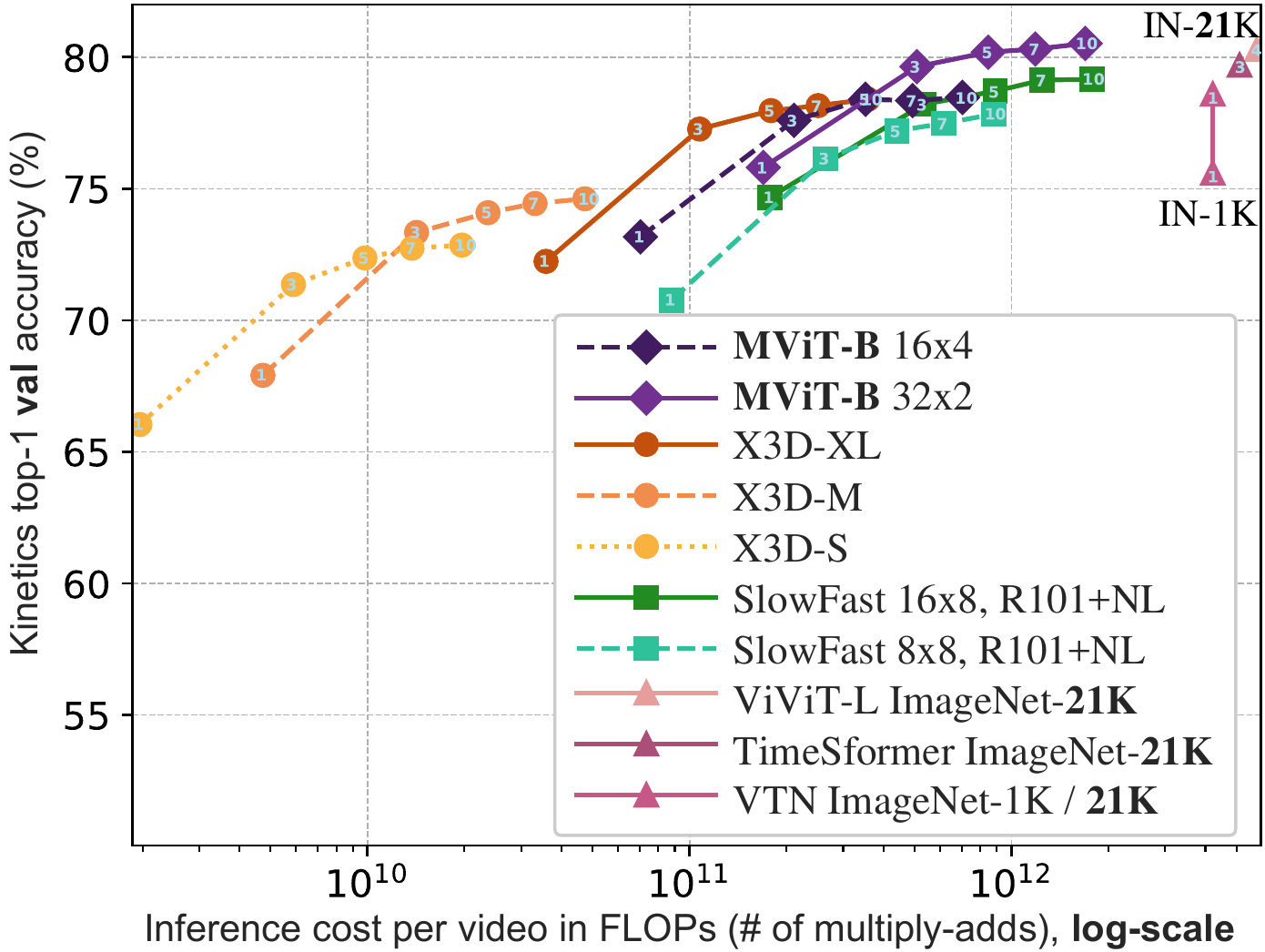}
	\caption{\textbf{Accuracy/complexity trade-off} on K400-\textbf{val} for varying  \# of inference clips per video. The top-1 accuracy (vertical axis) is obtained by \clipscolor{$K$}-Center clip testing where the number of temporal clips $\clipscolor{K}\in \{\clipscolor{1,3,5,7,10}\}$ is shown in each curve. The horizontal axis measures the full inference cost per video. The left-sided plots show a linear and the right plots a logarithmic (\textbf{log}) scale.
	}
	\label{fig:10clipTestAll}
\end{figure*}

\section{Additional Results} \label{sec:resultsapp}

\subsection{Ablations: Kinetics Action Classification}\label{sec:ablations_kinetics}

\paragraph{Inference cost.}

In the spirit of~\cite{feichtenhofer2020x3d} we aim to provide further ablations for the effect of using \textit{fewer }testing clips for efficient video-level inference. In \figref{fig:10clipTestAll} we analyze the trade-off for the full inference of a video, when varying the number of temporal clips used. The vertical axis shows the top-1 accuracy on K400-{val}  and the horizontal axis the overall inference cost in FLOPs for different  model families: \textbf{MViT}, X3D~\cite{feichtenhofer2020x3d}, SlowFast~\cite{Feichtenhofer2019}, and concurrent ViT models, VTN \cite{neimark2021video} ViT-B-TimeSformer \cite{bertasius2021space} ViT-L-ViViT \cite{arnab2021vivit}, pre-trained on ImageNet-\textbf{21K}. 

We first compare MViT with concurrent Transformer-based methods in the left plot in \figref{fig:10clipTestAll}. All these methods, VTN \cite{neimark2021video},  TimeSformer \cite{bertasius2021space} and ViViT \cite{arnab2021vivit}, pre-train on ImageNet-\textbf{21K} and use the ViT~\cite{dosovitskiy2020image} model with modifications on top of it. The inference FLOPs of these methods are around 5-10\x higher than MViT models with equivalent performance; for example, ViT-L-ViViT \cite{arnab2021vivit} uses 4 clips of 1446G FLOPs (\ie 5.78 TFLOPs) each to produce 80.3\% accuracy while MViT-B, 32\x3 uses 5 clips of 170G FLOPs (\ie 0.85 TFLOPs) to produce 80.2\% accuracy. Therefore, MViT-L can provide similar accuracy at 6.8\x~lower  FLOPs (and 8.5\x~lower parameters), than concurrent ViViT-L \cite{arnab2021vivit}. More importantly, the MViT result is achieved \textit{without external data}. All concurrent Transformer based works~\cite{neimark2021video,bertasius2021space,arnab2021vivit} require the huge scale ImageNet-21K to be competitive, and the performance degrades significantly (-3\% accuracy, see IN-1K in \figref{fig:10clipTestAll} for VTN \cite{neimark2021video}). These works further report failure of training without ImageNet initialization. 

The plot in~\figref{fig:10clipTestAll} right shows this same plot with a logarithmic scale applied to the FLOPs axis. Using this scaling it is clearer to observe that smaller models convolutional models ({X3D-S} and {X3D-M}) can still provide more efficient inference in terms of multiply-add operations and MViT-B compute/accuracy trade-off is similar to X3D-XL. 

\begin{figure}[h!]
	\vspace{-2pt}
	\centering
	\subfloat[\label{fig:qual:skip_connect1} \textbf{normalized} skip-connection]{\includegraphics[height = 0.47\linewidth]{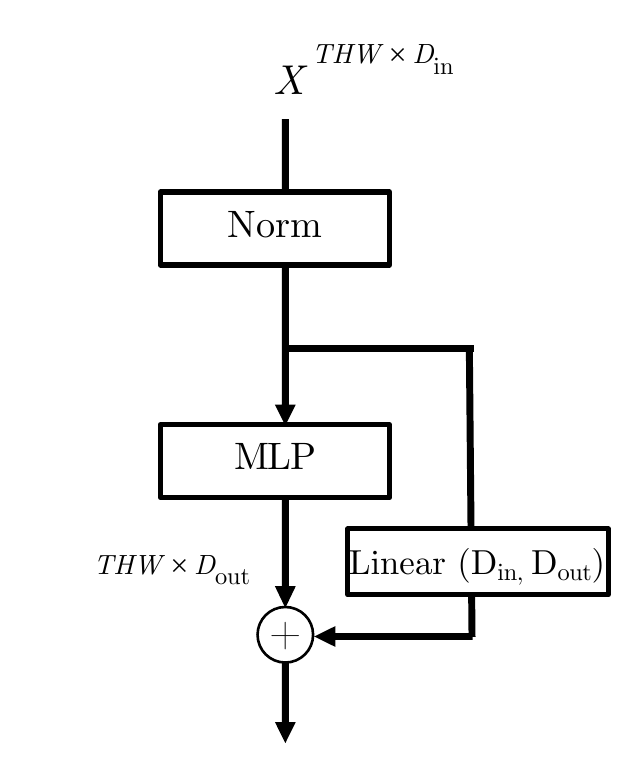}}
	\hfill
	\subfloat[\label{fig:qual:skip_connect2} unnormalized skip-connection]{\includegraphics[height = 0.47\linewidth]{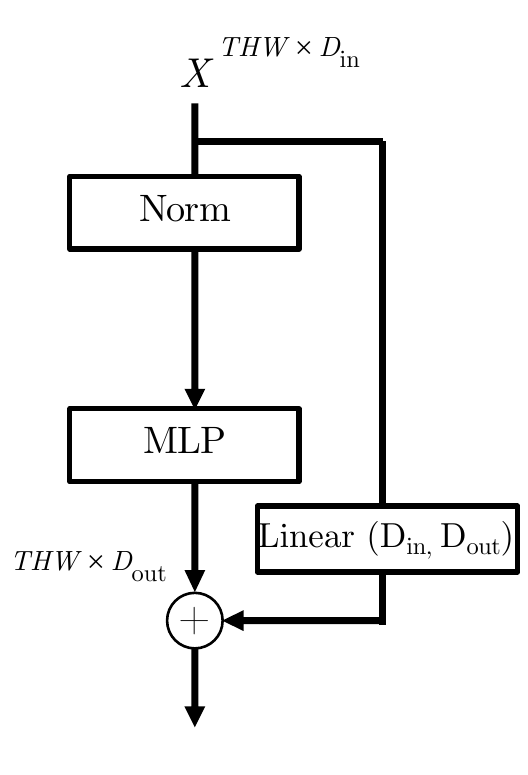}}
	\hfill
	\subfloat[\label{fig:qual:skip_connect3} no skip-connection]{\includegraphics[height = 0.47\linewidth]{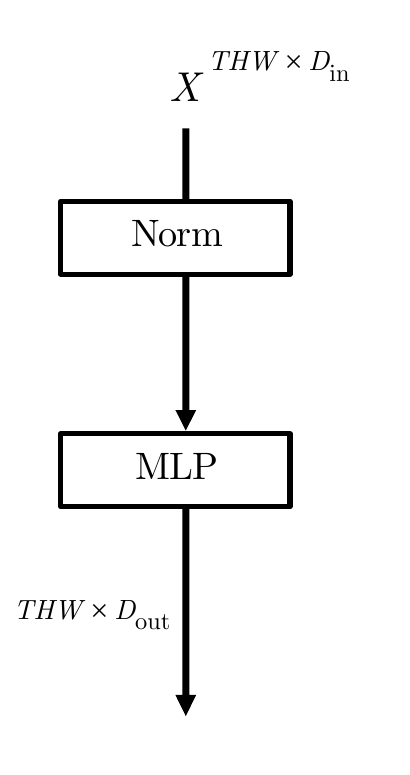}}   
	\vspace{5pt}
	\caption{\textbf{Skip-connections at stage-transitions.} Three skip-connection variants for expanding channel dimensions: (a) first normalize the input with layer normalization (Norm) and then expand its channel dimension; (b) directly expand the channel dimension of the input; (c) no skip-connection at stage-transitions.}
	\label{fig:qual:skip_connect}
\end{figure}

\paragraph{Ablations on skip-connections.} 
Recall that, at each scale-stage transition in MViT, we expand the channel dimension by increasing the output dimension of the previous stages' MLP layer; therefore, it is not possible to directly apply the original skip-connection design~\cite{dosovitskiy2020image}, because the input channel dimension ($D_\text{in}$) differs from the output channel dimension ($D_\text{out}$). We ablate three strategies for this:

(a) First normalize the input with layer normalization  and then expand its channel dimension to match the output dimension with a linear layer (Fig.~\ref{fig:qual:skip_connect1}); this is our default.  

(b) Directly expand the channel dimension of the input by using a linear layer to match the dimension (Fig.~\ref{fig:qual:skip_connect2}). 

(c) No skip-connection for stage-transitions (Fig.~\ref{fig:qual:skip_connect3}).

\begin{table}[h!]
	\centering
	\small
	\tablestyle{8pt}{1.1}
	\begin{tabular}{ll|c|x{22}}
		&	method   & top-1  & top-5  \\
		\shline
		(a) & \textbf{normalized} skip-connection & \textbf{77.2}  & \textbf{93.1} \\
		(b) & unnormalized skip-connection & 74.6 & 91.3 \\
		(c) & no skip-connection & 74.7 & 91.8 \\
		
	\end{tabular}
	\vspace{.5em}
	\caption{\textbf{Skip-connections at stage-transitions on K400.} We use our base model, {MViT}-B 16\x4. Normalizing the skip-connection at channel expansion is essential for good performance.}
	\label{tab:ablations:skip}
	
\end{table}

Table~\ref{tab:ablations:skip} shows the Kinetics-400 ablations for all 3 variants. Our default of using a normalized skip-connection  (a)  obtains the best results with 77.2\% top-1 accuracy, while using an un-normalized skip-connection after channel expansion (b) decays significantly to 74.6\% and using no skip-connection for all stage-transitions (c) has a similar result. We hypothesize that for expanding the channel dimension, normalizing the signal is essential to foster optimization, and use this design as our default in all other experiments. 

% slowfast with our recipe
\begin{table}[h]
	\centering
	\small
	\tablestyle{8pt}{1.1}
	\begin{tabular}{l|c|x{22}}
		backbone   & recipe  & Acc  \\
		\shline
		SlowFast R50, 8\x8 &  \cite{Feichtenhofer2019}  & 77.0 \\
		SlowFast R50, 8\x8 & MViT & 67.4 \\
		\hline
		SlowFast R101, 8\x8 &  \cite{Feichtenhofer2019} & 78.0 \\
		SlowFast R101, 8\x8 & MViT & 61.6 \\
	\end{tabular}
	\vspace{.5em}
	\caption{\textbf{SlowFast models with MViT recipe on Kinetics-400.} The default recipe is using the recipe from the original paper. Accuracy is evaluated on 10\x3 views. 
		\label{suptab:cnn_recipe}
	}
	\vspace{-1.2em}
\end{table}

\paragraph{SlowFast with MViT recipe.} To investigate if our training recipe can benefit ConvNet models, we apply the same augmentations and training recipe as for MViT to SlowFast in Table~\ref{suptab:cnn_recipe}. The results suggest that SlowFast models do not benefit from the MViT recipe directly and more studies are required to understand the effect of applying our training-from-scratch recipe to ConvNets, as it seems higher capacity ConvNets (R101) perform worse when using our recipe.

\subsection{Ablations: ImageNet Image Classification}\label{sec:ablations_imagenet}

We carry out ablations on ImageNet with the \textbf{MViT}-B-16 model with 16 layers, and show top-1 accuracy (Acc) as well as computational complexity measured in GFLOPs (floating-point operations). We also report Parameters in M($10^6$) and training GPU memory in G($10^9$) for a  batch size of 512.

\begin{table}[h]
	\centering
	\small
	\vspace{5pt}
	\tablestyle{3pt}{1.05}
	\begin{tabular}{c|x{26}|x{22}|x{22}}
		\multicolumn{1}{c|}{stride $\mathbf{s}$ }  & FLOPs  & Mem & Acc \\
		\shline
		8\x8 &  7.2 & 9.0  & 81.6 \\ 
		\underline{4\x4} & 7.8 & 11.9  & 82.5 \\ 
		2\x2 &  9.0 & 13.2 & 81.8 \\ 
		none &  10.4 & 17.3 & 82.3 \\ 
	\end{tabular}
	\vspace{.5em}
	\caption{ 
		\textbf{ImageNet: Key-Value pooling}: We vary stride  $ s_H \times s_W$, for pooling $K$ and $V$. We use ``adaptive'' pooling that reduces stride \wrt stage resolution.  \label{tab:ablation:IN:kv_pool}}
\end{table}

\paragraph{Key-Value pooling for image classification.} The ablation in \tblref{tab:ablation:IN:kv_pool} analyzes the pooling stride  $\mathbf{s}=s_H \times s_W$, for pooling $K$ and $V$ tensors. Here, we use our default `adaptive' pooling that uses a stride \wrt stage resolution, and keeps the $K,V$ resolution \textit{fixed} across all stages. 

First, we compare the baseline which uses pooling with a fixed stride of 4\x4 with a model has a stride of 8\x8: this drops accuracy from 82.5\% to 81.6\%, and reduces FLOPs and memory by 0.6G and 2.9G. 

Second, we reduce the stride to 2\x2, which increases FLOPs and memory significantly but performs 0.7\% \textit{worse} than our default stride of  4\x4.

Third, we remove the  $K,V$ pooling completely which increases FLOPs by 33\% and memory consumption by 45\%, while providing lower accuracy than our default.

Overall, the results show that our $K,V$ pooling is an effective technique to \textit{increase} accuracy \textit{and} \textit{decrease} cost (FLOPs/memory) for image classification.

\begin{figure*}[t!]
	\vspace{-10pt}
	\centering
	
	\subfloat[\label{fig:qual:mad_init_vit} ViT-B at \textbf{initialization}]{\includegraphics[width = 0.47\linewidth]{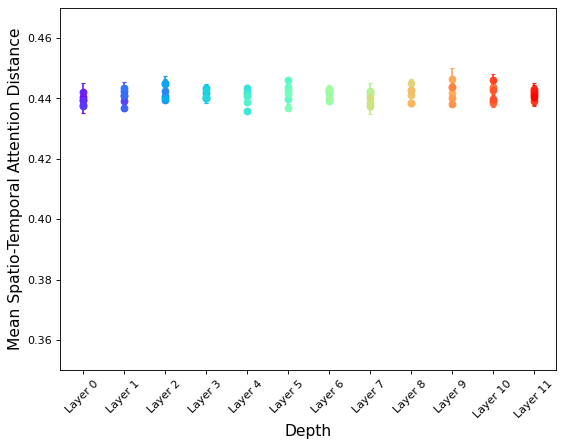} }%
	\hfill
	\subfloat[\label{fig:qual:mad_init_mvit} MViT-B at \textbf{initialization}]{\includegraphics[width = 0.47\linewidth]{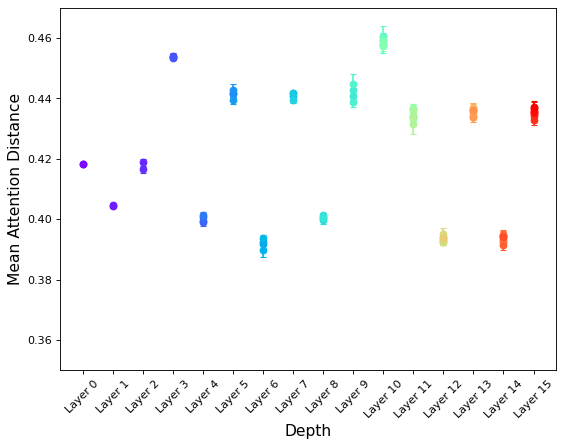} }%
	\hfill
	\subfloat[\label{fig:qual:mad_converged_vit} ViT-B at \textbf{convergence}]{\includegraphics[width = 0.47\linewidth]{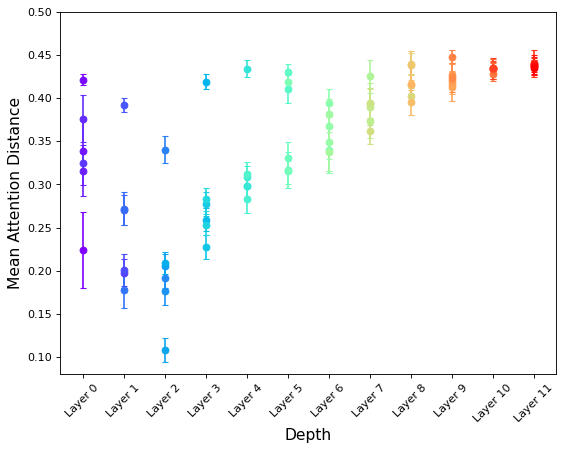} }%
	\hfill
	\subfloat[\label{fig:qual:mad_converged_mvit} MViT-B at \textbf{convergence}]{\includegraphics[width = 0.47\linewidth]{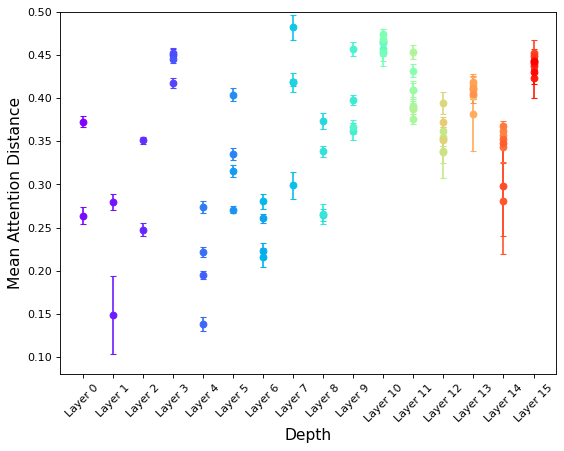} }%
	\hfill 
	
	\vspace{5pt}
	\caption{\textbf{Mean attention distance} across layers \textbf{\textit{at initialization/convergence}} for Vision Transformer \protect\subref{fig:qual:mad_init_vit}/\protect\subref{fig:qual:mad_converged_vit} \& Multiscale Vision Transformers \protect\subref{fig:qual:mad_init_mvit}/\protect\subref{fig:qual:mad_converged_mvit}. Each point shows the normalized average attention distance (weighted by the attention scores, with 1.0 being maximum possible distance) for each head in a layer. MViT attends close and distant features throughout the network hierarchy.}
	\label{fig:qual:mad}
	\vspace{-5pt}
\end{figure*}

\section{Qualitative Experiments: Kinetics}\label{sec:ablations_qualitative}

% \paragraph{Qualitative Experiments.} 
In Figure \ref{fig:qual:mad}, we plot the mean attention distance for all heads across all the layers of our Multiscale Transformer model and its Vision Transformer counterpart, at initialization with random weights, and at convergence after training. Each head represents a point in the plots (ViT-B has more heads). Both the models use the exact same weight initialization scheme and the difference in the attention signature stems purely from the multiscale skeleton in MViT. 
We observe that the dynamic range of attention distance is about \textbf{4\x} larger in the MViT model than ViT \textit{at initialization} itself (\ref{fig:qual:mad_init_vit} \vs  \ref{fig:qual:mad_init_mvit}). This signals the strong inductive bias stemming from the multiscale design of MViT. Also note that while at initialization, every layer in ViT has roughly the same mean attention distance, the MViT layers have strikingly different mean attention signatures indicating distinct predilections towards global and local features. 

The bottom row of \figref{fig:qual:mad} shows the same plot for a converged Vision Transformer (\ref{fig:qual:mad_converged_vit}) and Multiscale Vision Transformer  (\ref{fig:qual:mad_converged_mvit}) model. 

We notice very different trends between the two models\textit{ after training}. While the ViT model  (\ref{fig:qual:mad_converged_vit}) has a consistent increase in attention distance across layers, the MViT model  (\ref{fig:qual:mad_converged_mvit}) is not monotonic at all. Further, the intra-head variation in the ViT model decreases as the depth saturates, while, for MViT, different heads are still focusing on different features even in the higher layers. This suggests that some of the capacity in the ViT model might indeed be wasted with redundant computation while the lean MViT heads are more judiciously utilizing their compute. Noticeable is further a larger delta (between initialization in \figref{fig:qual:mad_init_vit} and convergence in \ref{fig:qual:mad_converged_vit}) in the overall attention distance signature in the ViT model, compared to MViT's location distribution. 

\section{Computational Analysis}
\label{sec:analysis}
Since attention is quadratic in compute and memory complexity, pooling the key, query and value vectors have direct benefits on the fundamental compute and memory requirements of the pooling operator and by extension, on the complete \fullnameCC model. Consider an input tensor of dimensions $T \times H \times W$ and  corresponding sequence length $L = T \cdot H \cdot W$. Further, assume the key, query and value strides to be $\mathbf{s}^K$, $\mathbf{s}^Q$ and $\mathbf{s}^V$. As described in 
Sec.~{\color{red}3.1} in main paper,
each of the vectors would experience a sptio-temporal resolution downsampling by a factor of their corresponding strides. Equivalently, the sequence length of query, key and value vectors would be reduced by a factor of $f^Q$, $f^K$ and $f^V$ respectively, where,

$$
f^j = s^j_T \cdot s^j_H \cdot s^j_W, \ \forall \ j \in \{Q,K,V\}.
$$
\vspace{-10pt}
\paragraph{Computational complexity.} Using these shorter sequences yields a corresponding reduction in space and runtime complexities for the pooling attention operator. Considering key, query and value vectors to have sequence lengths $L/f_k$, $L/f_q$ and $L/f_v$ after pooling, the overall runtime complexity of computing the key, query and value embeddings is $O(THWD^2/h)$ per head, where $h$ is the number of heads in \attnabbvspace. Further, the runtime complexity for calculating the full attention matrix and the weighed sum of value vectors with reduced sequence lengths is $O(T^2H^2W^2D/f_qf_hh)$ per head. Computational complexity for pooling is 

$$T(\mathcal{P}(\cdot ; \mathbf{\Theta})) =  O\left(THW \cdot D \cdot \frac{k_Tk_Wk_H}{s_Ts_Ws_H}\right),$$ 

which is negligible compared to the quadratic complexity of the attention computation and hence can be ignored in asymptotic notation. Thus, the final runtime complexity of \attnabbvspace is $O(THWD(D + THW/f_qf_k))$. 

\paragraph{Memory complexity.} The space complexity for storing the sequence itself and other tensors of similar sizes is  $O(THWD)$. Complexity for storing the full attention matrix is $O(T^2H^2W^2 h/f_qf_k)$. Thus the total space complexity of \attnabbvspace is $O(THWh (D/h + THW/f_qf_k))$. 

\paragraph{Design choice.} Note the trade-off between the number of channels $D$ and the sequence length term $THW/f_qf_k$ in both space and runtime complexity. This tradeoff in \attnname informs two critical design choices of \fullnameCC architecture. 

First, as the effective spatio-temporal resolution decreases with layers because of diminishing $THW/f_qf_k$, the channel capacity is increased to keep the computational time spent (FLOPs) roughly the same for each stage. 

Second, for a fixed channel dimension, $D$, higher number of heads $h$ cause a prohibitively larger memory requirement because of the $(D + h * THW/f_qf_k)$ term. Hence, \fullnameCC starts with a small number of heads which is increased as the resolution factor $THW/f_qf_k$ decreases, to hold the effect of $(D + h * THW/f_qf_k)$ roughly constant.

\section{Additional Implementation Details} \label{sec:app_training} 

We implement our model with \mbox{PySlowFast}~\cite{fan2020pyslowfast}. Code and models are available at: \url{https://github.com/facebookresearch/SlowFast}. 

\subsection{Details: Kinetics Action Classification}\label{sec:kineticsapp}
\paragraph{Architecture details.}
As in original ViT~\cite{dosovitskiy2020image}, we use residual connections~\cite{He2016} and Layer Normalization (LN) \cite{ba2016layer} in the pre-normalization configuration that applies LN at the beginning of the residual function, and our MLPs consist of two linear layers with GELU activation~\cite{hendrycks2016gaussian}, where the first layer expands the dimension from $D$ to $4D$, and the second restores the input dimension $D$, except at the end of a scale-stage, where we increase this channel dimensions to match the input of the next scale-stage. At such stage-transitions, our skip connections receive an extra linear layer that takes as input the layer-normalized signal which is also fed into the MLP. In case of $Q$-pooling at scale-stage transitions, we correspondingly pool the skip-connection signal.

\paragraph{Optimization details.}
We use the truncated normal distribution initialization in~\cite{hanin2018start} and adopt synchronized AdamW~\cite{loshchilov2018fixing} training on 128 GPUs following the recipe in \cite{touvron2020training,Feichtenhofer2019}.
For Kinetics, we train for 200 epochs with 2 repeated augmentation~\cite{hoffer2020augment} repetitions. The mini-batch size is 4 clips per GPU (so the overall $\mathrm{batchsize}$ is $512$).

We adopt a half-period cosine schedule \cite{Loshchilov2016} of learning rate decaying: the learning rate at the $n$-th iteration is $\eta\cdot0.5[\cos(\frac{n}{n_\text{max}}\pi)+1]$, where $n_\text{max}$ is the maximum training iterations and the base learning rate $\eta$ is set as $1.6\cdot10^{-3}$. We linearly scale the base learning rate \wrt the overall batch-size, $\eta =1.6{\cdot}10^{-3}\frac{\mathrm{batchsize}}{512}$, and use a linear warm-up strategy in the first 30 epochs \cite{Goyal2017}. The cosine schedule is completed when reaching a final learning rate of $1.6\cdot10^{-5}$. We extract the class token after the last stage and use it as the input to the final linear layer to predict the output classes.  
For \textbf{Kinetics-600} all hyper-parameters are identical to K400.  

\paragraph{Regularization details.}
We use weight decay of 5$\cdot$10$^\text{-2}$, a dropout \cite{Hinton2012b} of 0.5 before the final classifier, label-smoothing~\cite{Szegedy2015a} of 0.1 and  use stochastic depth~ \cite{huang2016deep} (\ie drop-connect) with rate 0.2.

Our data augmentation is performed on input clips by applying the same transformation across all frames. 
To each clip, we apply a random horizontal flip, 
Mixup \cite{zhang2018mixup} with $\alpha=0.8$ to half of the clips in a batch and CutMix~\cite{yun2019cutmix} to the other half, Random Erasing~\cite{zhong2020random} with probability $0.25$, and 
Rand Augment \cite{cubuk2020randaugment} with probability of $0.5$ for $4$ layers of maximum magnitude $7$.

For the temporal domain, we randomly sample a clip from the full-length video, and the input to the network are  $T$~frames with a temporal stride of $\tau$; denoted as $T\times\tau$~\cite{Feichtenhofer2019}. For the spatial domain, we use  Inception-style \cite{Szegedy15} cropping that randomly resizes the input \textit{area} between a $[$min, max$]$, scale of $[$0.08,  1.00$]$, and jitters aspect ratio between 3/4 to 4/3, before taking an $H\times W$ = 224\x224 crop. 

\paragraph{Fine-tuning from ImageNet.} 

To fine-tune our ViT-B baseline, we extend it to take a video clip of $T=8$ frames as input and initialize the model weights from the ViT-B model~\cite{dosovitskiy2020image} pre-trained on ImageNet-21K dataset. The positional embedding is duplicated for each frame. We fine-tune the model for 30 epochs with SGD using the recipe in ~\cite{Feichtenhofer2019}. 
The mini-batch size is 2 clips per GPU and a half-period cosine learning rate decay is used. We linearly scale the base learning rate \wrt the overall batch-size, $\eta =10^{-3}\frac{\mathrm{batchsize}}{16}$. Weight decay is set to $10^{-4}$.

\subsection{Details: AVA Action Detection}\label{sec:detection}

\paragraph{Dataset.}
The AVA dataset \cite{Gu2018} has bounding box annotations for spatiotemporal localization of (possibly multiple) human actions. It has 211k training and 57k validation video segments. We follow the standard protocol reporting mean Average Precision (mAP) on 60 classes \cite{Gu2018} on AVA v2.2.

\paragraph{Detection architecture.}
We follow the detection architecture in \cite{Feichtenhofer2019} to allow direct comparison of MViT against \mbox{SlowFast} networks as a backbone.

First, we reinterpret our transformer spacetime cube outputs from MViT as a spatial-temporal feature map by concatenating them according to the corresponding temporal and spatial location. 

Second, we employ a the detector similar to Faster R-CNN \cite{Ren2015} with minimal modifications adapted for video. 
Region-of-interest (RoI) features \cite{Girshick2015} are extracted at the generated feature map from MViT by extending a 2D proposal at a frame into a 3D RoI by replicating it along the temporal axis, similar as done in previous work \cite{Gu2018,Sun2018,Jiang2018}, followed by application of frame-wise RoIAlign \cite{He2017} and temporal global average pooling. The RoI features are then max-pooled and fed to a per-class, sigmoid classifier for prediction. 

\paragraph{Training.} 
We initialize the network weights from the Kinetics models and adopt synchronized SGD training on 64 GPUs. We use 8 clips per GPU as the mini-batch size and a half-period cosine schedule of learning rate decaying. The base learning rate is set as $0.6$.
We train for 30 epochs with linear warm-up \cite{Goyal2017} for the first 5 epochs and use a weight decay of 10$^{-8}$ and stochastic depth~\cite{huang2016deep} with rate 0.4.
Ground-truth boxes, and proposals overlapping with ground-truth boxes by \mbox{IoU $>$ 0.9}, are used as the samples for training. 
The region proposals are identical to the ones used in \cite{Feichtenhofer2019}.

\paragraph{Inference.} We perform inference on a single clip with $T$~frames sampled with stride $\tau$~centered at the frame that is to be evaluated.

\subsection{Details: Charades Action Classification}\label{sec:charades}
\paragraph{Dataset.}
Charades \cite{Sigurdsson2016} has $\app$9.8k training videos and 1.8k validation videos in 157 classes in a multi-label classification setting of longer activities spanning $\app$30 seconds on average. Performance is measured in mean Average Precision (mAP).

\paragraph{Training.} We fine-tune our MViT models from the Kinetics models.  A per-class sigmoid output is used to account for the multi-class nature. We train with SGD on 32 GPUs for 200 epochs using 8 clips per GPU. The base learning rate is set as 0.6 with half-period cosine decay. We use weight decay of 10$^\text{-7}$ and stochastic depth~\cite{huang2016deep} with rate 0.45. We perform the same data augmentation schemes as for Kinetics in \S\ref{sec:kineticsapp}, except of using Mixup.

\paragraph{Inference.} 
To infer the actions over a single video, we spatio-temporally max-pool prediction scores from multiple clips in testing \cite{Feichtenhofer2019}.

\subsection{Details: Something-Something V2 (SSv2)}\label{sec:ssv2}

\paragraph{Dataset.}
The Something-Something V2 dataset~\cite{ssv2} contains 169k training, and 25k validation videos.
The videos show human-object interactions to be classified into 174 classes.
We report accuracy on the validation set. 

\paragraph{Training.}
We fine-tune the pre-trained Kinetics models. We train for 100 epochs using 64 GPUs with 8 clips per GPU and a base learning rate of 0.02 with half-period cosine decay \cite{Loshchilov2016}. Weight decay is set to 10$^{-4}$ and stochastic depth rate~\cite{huang2016deep} is 0.4. Our training augmentation is the same as in \S\ref{sec:kineticsapp}, but as SSv2 requires distinguishing between directions, we disable random flipping in training.
We use segment-based input frame sampling~\cite{lin2018temporal} that splits each video into segments, and from each of them, we sample one frame to form a clip.

\paragraph{Inference.} We take single clip with 3 spatial crops to form predictions over a single video in testing.

\subsection{Details: ImageNet}\label{sec:details_imagenet}

\paragraph{Datasets.} For image classification experiments, we perform our experiments on ImageNet-1K~\cite{Deng2009} dataset that has $\app$1.28M images in 1000 classes.
We train models on the {train} set and report top-1 and top-5 classification accuracy (\%) on the  {val} set. Inference cost (in FLOPs) is measured from a single center-crop with resolution of $224^2$ if the input resolution was not specifically mentioned.

\paragraph{Training.}
We use the training recipe of DeiT~\cite{touvron2020training} and summarize it here for completeness. We train for $100$ epochs with $3$ repeated augmentation~\cite{hoffer2020augment} repetitions (overall computation equals $300$ epochs), using a batch size of $4096$ in $64$ GPUs. We use truncated normal distribution initialization~\cite{hanin2018start} and adopt synchronized AdamW~\cite{loshchilov2018fixing} optimization with a base learning rate of $0.0005$ per $512$ batch-size that is warmed up and decayed as half-period cosine, as in~\cite{touvron2020training}. We use a weight decay of $0.05$, label-smoothing~\cite{Szegedy2015a} of $0.1$. Stochastic depth~ \cite{huang2016deep} (\ie drop-connect) is also used with rate $0.1$ for model with depth of 16 ({MViT}-B-16), and rate $0.3$ for deeper models ({MViT}-B-24). Mixup \cite{zhang2018mixup} with $\alpha=0.8$ to half of the clips in a batch and CutMix~\cite{yun2019cutmix} to the other half, Random Erasing~\cite{zhong2020random} with probability $0.25$, and 
Rand Augment \cite{cubuk2020randaugment} with maximum magnitude $9$ and probability of $0.5$ for $4$ layers (for max-pooling) or $6$ layers (for conv-pooling). 
\section*{Acknowledgements} We are grateful for discussions with Chao-Yuan Wu, Ross Girshick, and Kaiming He.\\ 
{
	\small
	\bibliographystyle{ieee_fullname}
	\bibliography{mvit}
}

\end{document}